\documentclass[a4paper,fleqn]{cas-sc}

\usepackage[authoryear,longnamesfirst]{natbib}
\usepackage{graphicx}
\usepackage{float}
\usepackage{color, xcolor}
\usepackage{subfigure}
\usepackage{caption}
\usepackage{amsmath,amssymb}
\usepackage{booktabs}
\usepackage{multirow}
\usepackage{array}
\usepackage[misc]{ifsym}
\usepackage{threeparttable}
\usepackage[normalem]{ulem}
\usepackage{url}

\usepackage{hyperref}

\usepackage[capitalize]{cleveref}
\usepackage{lineno} 
\usepackage{latexsym}
\usepackage{amssymb}
\usepackage{amsmath}
\usepackage{amsthm}
\usepackage{booktabs}
\usepackage{enumitem}
\usepackage{graphicx}
\usepackage{color}

\usepackage{multirow}
\usepackage{multicol}
\usepackage{subfigure}
\usepackage{bbding}
\usepackage{amssymb}
\usepackage{makecell}
\usepackage{pifont}
\usepackage{tcolorbox}
\usepackage{url}
\usepackage{cleveref}
\usepackage{algorithm}
\usepackage{algpseudocode}

\crefname{figure}{Fig.}{Fig.}

\newdefinition{definition}{Definition}

\def\tsc#1{\csdef{#1}{\textsc{\lowercase{#1}}\xspace}}
\tsc{WGM}
\tsc{QE}
\tsc{EP}
\tsc{PMS}
\tsc{BEC}
\tsc{DE}
\begin{document}
\let\WriteBookmarks\relax
\def\floatpagepagefraction{1}
\def\textpagefraction{.001}

\shorttitle{Adaptive-Solver Framework for Dynamic Strategy Selection in Large Language Model Reasoning}

\shortauthors{Jianpeng Zhou et~al.}

\title [mode = title]{Adaptive-Solver Framework for Dynamic Strategy Selection in Large Language Model Reasoning}                      
% \tnotemark[1,2]

\author[1]{Jianpeng Zhou}[style=chinese]
% \cormark[2]
\ead{zhoujp7@mail2.sysu.edu.cn}
\credit{Conceptualization, Methodology, Software, Writing - Original draft}
\fnmark[1]

\author[1]{Wanjun Zhong}[style=chinese]
\ead{zhongwj25@mail2.sysu.edu.cn}
\credit{Conceptualization, Writing- Reviewing and Editing}
\fnmark[1]

\author[2]{Yanlin Wang}[style=chinese]
\ead{wangylin36@mail.sysu.edu.cn}
\credit{Supervision, Writing- Reviewing and Editing}

\author[1]{Jiahai Wang}[style=chinese]
\ead{wangjiah@mail.sysu.edu.cn}
\credit{Supervision, Writing- Reviewing and Editing, Funding acquisition}
\cormark[1]

% % Footnote of the first author

\affiliation[1]{organization={School of Computer Science and Engineering},
    addressline={Sun Yat-sen University}, 
    city={Guangzhou},
    citysep={}, % Uncomment if no comma needed between city and postcode
    postcode={510006}, 
    country={China}}
% \ead{cvr3@sayahna.org}
% \ead[URL]{www.sayahna.org}

\affiliation[2]{organization={School of Software Engineering},
    addressline={Sun Yat-sen University}, 
    city={Zhuhai},
    citysep={}, % Uncomment if no comma needed between city and postcode
    postcode={519000}, 
    % state={Trivandrum},
    country={China}}

\cortext[cor1]{Corresponding author}
% \cortext[cor2]{Principal corresponding author}

\fntext[fn1]{Equal contribution.}

\let\WriteBookmarks\relax
\let\printorcid\relax    % 可去掉页面下方的ORCID(s)


\begin{abstract}
Large Language Models (LLMs) demonstrate impressive ability in handling reasoning tasks.
However, unlike humans who can instinctively adapt their problem-solving strategies to the complexity of task, most LLM-based methods adopt a one-size-fits-all approach. These methods employ consistent models, sample sizes, prompting methods and levels of problem decomposition, regardless of the problem complexity.
The inflexibility of these methods can bring unnecessary computational overhead or sub-optimal performance. To address this limitation, we introduce an Adaptive-Solver (AS) framework that \textbf{dynamically adapts solving strategies to suit various problems, enabling the flexible allocation of test-time computational resources.}
The framework functions with two primary modules. 
The initial \textit{evaluation} module assesses the reliability of the current solution using answer consistency. If the solution is deemed unreliable, the subsequent \textit{adaptation} module comes into play. 
Within this module, various types of adaptation strategies are employed collaboratively. Through such dynamic and multi-faceted adaptations, our framework can help reduce computational consumption and improve performance. 
Experimental results from complex reasoning benchmarks reveal that our method can significantly reduce API costs (up to 85\%) while maintaining original performance. Alternatively, it achieves up to 4.5\% higher accuracy compared to the baselines at the same cost. The code and dataset are available at https://github.com/john1226966735/Adaptive-Solver.

\end{abstract}

% Keywords

\maketitle

\section{Introduction}

Large Language Models (LLMs) have demonstrated significant potential across various reasoning tasks~\citep{survey-llm-reason,survey-llm-prompt-reason}. 
However, while their ability to tackle complex problems is evident, an optimized strategy that balances maximizing performance with minimizing resource consumption remains underexplored. Identifying such an effective problem-solving approach is critical yet challenging.
To address this challenge, we draw inspiration from human problem-solving techniques.
The human cognitive framework consists of two distinct systems: \textit{System 1} for intuitive thinking, and \textit{System 2} for deeper, analytical reasoning~\citep{sloman1996empirical,daniel2017thinking}. These systems are utilized dynamically and adaptably, catering to a range of problem complexities, thereby ensuring both efficiency and accuracy in problem-solving.

Likewise, when facing complex challenges, humans often break down the problem into simpler sub-questions, ensuring a lucid formulation of the task. For simpler question, a direct, singular line of reasoning is typically employed. 
If their initial solution does not meet expectations, humans naturally pivot their approach in pursuit of a more effective resolution. 
Inspired by these flexible human strategies, we propose that machines, specifically LLMs, should be equipped with similar adaptability.
This adaptation may involve adjusting various aspects of the solving strategy, such as the LLM models, sample sizes, prompting techniques, and the granularity of problem decomposition.

\begin{figure}[h]
    \centering
    \includegraphics[width=0.8\linewidth]{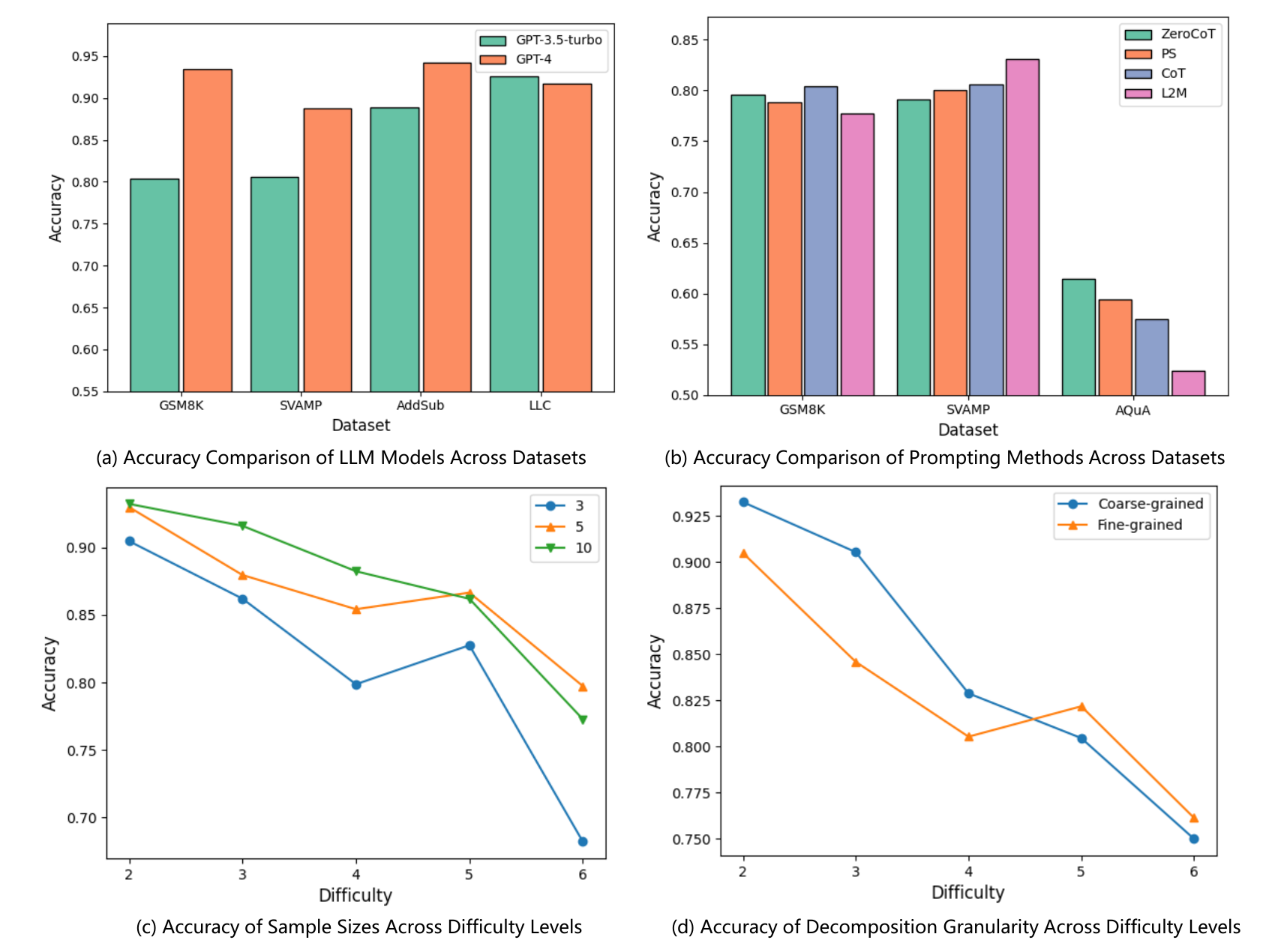}
    \caption{
    A motivation illustration. 
    \textit{Difficulty} is measured by the number of steps in the ground-truth solution.
    (a) The performance advantage of a larger, more expensive model over a smaller, cheaper model varies across datasets; for simpler tasks, smaller models can perform comparably to larger ones. (b) Different prompting methods have unique strengths, so the optimal prompting approach depends on the characteristics of each dataset. (c) For tasks of varying difficulty, particularly simpler ones, using a smaller sample size can achieve similar accuracy as a larger sample size while reducing costs. (d) For tasks with different difficulty levels, the ideal decomposition granularity varies.
    }
    \label{fig:motivation_illustration}
\end{figure}

To analyze how different LLM models, sample sizes, prompting methods and decomposition granularity perform across datasets or task difficulties, we conducted a series of ablation experiments on existing methods. Our findings reveal that the optimal balance between performance and cost often varies depending on the dataset or task difficulty.

As shown in Figure 1(a), although GPT-4 generally outperforms GPT-3.5-turbo, simpler datasets allow the cheaper GPT-3.5-turbo model to perform comparably to GPT-4. This suggests that dynamically selecting a smaller model for less complex tasks could reduce costs without significantly sacrificing accuracy. 
At the problem-solving method layer, several prompting methods, including CoT~\citep{CoT}, ZeroCoT~\citep{ZeroCoT}, L2M~\citep{L2M}, and PS~\citep{Plan-and-solve}, have been proposed to instruct LLMs to follow specific problem-solving strategies. As shown in Figure 1(c), each method has its own strengths and performs differently across datasets. This underscores the importance of choosing the right prompting technique for each dataset to achieve the best results.
Self-Consistency (SC)~\citep{SC} improves CoT by exploring multiple reasoning paths and selecting the most consistent answer, which helps reduce the internal randomness of LLMs. While increasing the sample size (i.e., the number of reasoning paths) generally enhances accuracy, it also raises computational costs. Figure 1(b) shows that for simpler problems, a moderate sample size (e.g., 5) performs similarly to a larger sample size (e.g., 10). This suggests that adjusting sample sizes based on problem difficulty can help balance cost and performance.
For complex, multi-step problems, L2M prompting decomposes the original question into simpler sub-questions, solving each sequentially to arrive at the final answer. As shown in Figure 1(d), the granularity of decomposition impacts performance: coarse-grained decomposition (fewer sub-questions) works better for simpler tasks, whereas fine-grained decomposition (more sub-questions) enhances accuracy for more complex tasks. This finding implies that adapting decomposition granularity to problem difficulty is critical for maximizing the effectiveness of decomposition-based prompting techniques like L2M.

Most current approaches rely on static solvers\footnote{In this context, a \textbf{solver} encompasses all elements integral to problem-solving, including the LLM model, sample size, prompting technique, decomposition strategy, and so forth.}, overlooking the unique characteristics of individual problems. 
This rigidity can lead to unnecessary resource consumption and suboptimal performance due to the fixed allocation of computational resources. 
Therefore, we argue that \textbf{dynamically customized solvers are essential for achieving both cost-efficiency and enhanced performance across diverse tasks}.
Recent studies suggest that scaling test-time computation can effectively enhance model performance~\citep{scaling_test_time,empirical}. Complex problems should be allocated more computational resources during testing. However, determining how to best allocate computational resources across different problems is a question that warrants further study.

\begin{figure}[h]
\begin{center}
\includegraphics[width=0.8\linewidth]{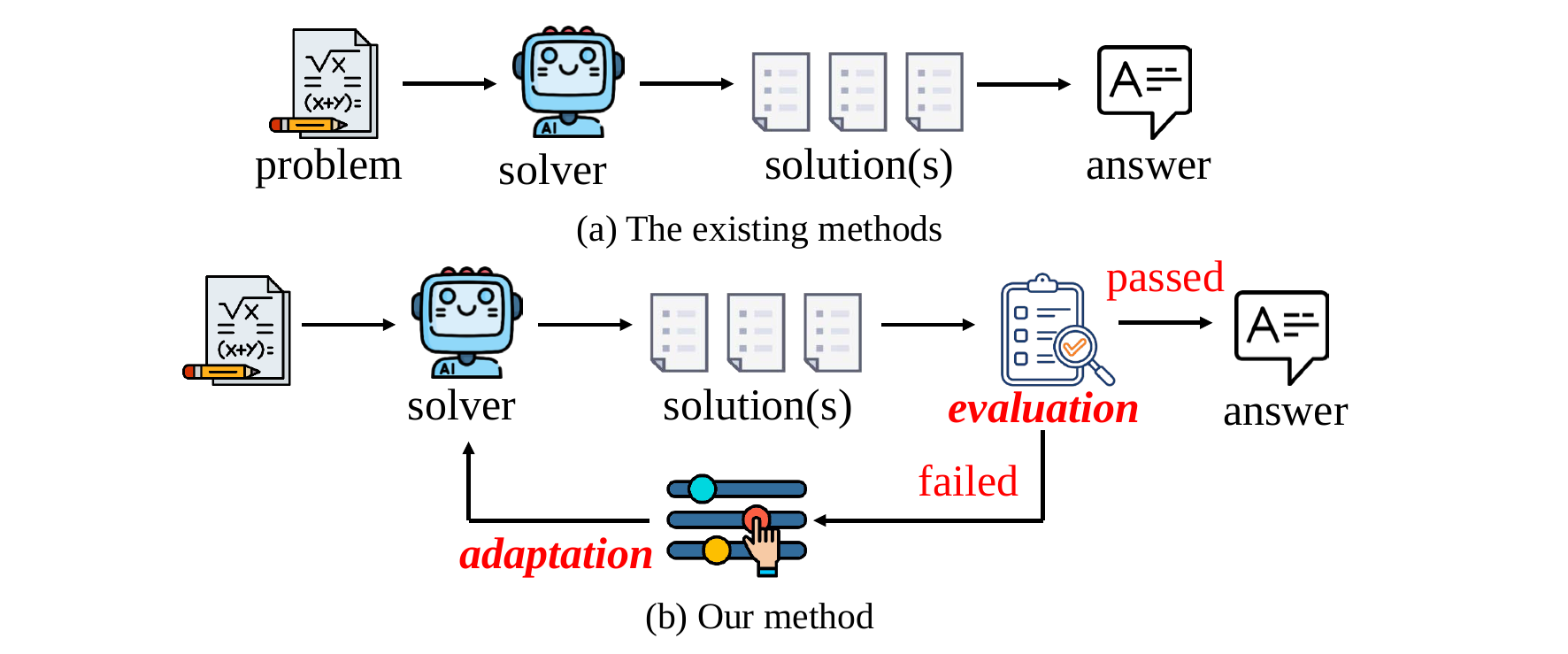}
\end{center}
\caption{Comparison of the frameworks of our method and baselines. (a) Existing methods utilize static solvers. (b) Our framework selects a suitable solver from candidate solvers for each different problem.
The \textcolor{red}{red} section highlights the differences between our method and the baselines.
}
\label{motivation}
\end{figure}

In response to the clear demand for dynamic problem-solving methods, we propose the Adaptive-Solver (AS) framework, as illustrated in Figure~\ref{motivation}(b). 
The AS framework consists of two primary components: the \textit{evaluation} module and the \textit{adaptation} module. The \textit{evaluation} module assesses the current solution's quality, determining whether the problem has been adequately solved. If the solution falls short, the \textit{adaptation} module is triggered to adjust the solving strategy in the subsequent round.
Within the \textit{adaptation} module, four adaptation strategies are devised: 
(1) \textit{Model Adaptation}: Shifting to a more powerful, albeit resource-intensive, LLM when necessary;
(2) \textit{Sample Size Adaptation}: Initializing the sample size with small value and incrementally lifting it when needed;
(3) \textit{Prompting Method Adaptation}: Varying the prompting techniques to better align with the complexity of the problem;
(4) \textit{Decomposition Granularity Adaptation}: Modulating the granularity of problem decomposition according to the problem complexity.
These adaptation strategies can be combined to achieve a dynamic and multifaceted adjustment to the current solver.
This flexible adjustment enables our method to address the issues of unnecessary resource consumption and suboptimal performance in existing approaches by adaptively selecting the most effective and cost-efficient solver for each task.

Extensive experiments across 8 reasoning tasks corroborate the effectiveness of the Adaptive-Solver and draw several crucial findings:
1) Compared with using the most powerful model (i.e., GPT4) alone, our method achieves a 46\%-85\% reduction in inference costs while maintaining comparable performance. 2) At equivalent costs, our method demonstrates superior performance than other baselines. These results show that our method offers the dual advantage of reducing costs while improving performance.

\paragraph{Contributions.} 
Our key contributions are as follows: 
\begin{itemize}
\item \textbf{Framework.} 
We introduce the Adaptive-Solver framework, which dynamically adjusts inference strategies from multiple aspects, including LLM models, the number of solving attempts, prompting methods, and decomposition granularity, based on the difficulty of the given problem. This allows for the flexible combination of different inference strategies, resulting in a better cost-effectiveness trade-off.

\item \textbf{Algorithms.} We propose four versatile adaptation strategies concerning the selection of LLM model, sample size, prompting method and decomposition granularity. Furthermore, we devise an algorithm that integrates these strategies to facilitate efficient adjustment of solver.

\item \textbf{Experiments.} Experiments underscore the superiority of the Adaptive-Solver framework, demonstrating marked enhancements in computational efficiency and performance outcomes.
\end{itemize}

\section{Related Work}
\label{Related work}

\subsection{Deep Learning for Math Word Problems}
Designing algorithms to automatically solve math word problems (MWPs) has long been a focus of NLP research. With the rise of large language models (LLMs), there has been an increasing interest in leveraging LLMs for MWP solving. Before the advent of LLMs, several types of deep learning approaches were proposed for MWPs. These can be broadly categorized into four types~\citep{survey-dl-math}: Seq2Seq-based, graph-based, attention-based, and pre-trained language model-based. 
1) \textbf{Seq2Seq-based models}~\citep{DNS,program-induction} are the first to apply deep learning to MWPs, leveraging an encoder-decoder architecture typically modeled by Recurrent Neural Networks. The key idea is to map a math problem description into a mathematical expression or equation, which is then solved by a symbolic solver. However, these models ignore the structural information inherent in math problems or mathematical expressions, which can be represented as trees or graphs.
2) To address this, \textbf{graph-based methods} explicitly incorporate the structure of math problems or expressions in the encoder or decoder. For instance, Sequence-to-tree models~\citep{GTS,KA-S2T} explicitly model the tree structure when encoding output sequences. 
NERHRT~\citep{NERHRT} introduces a hierarchical recursive tree-structured decoder to mitigate the early information loss in the tree decoder.
Graph-to-tree models utilize graph encoders to embed structural information from math problems. 
For example, Graph2Tree-Z~\citep{g2t-z} constructs quantity cell and quantity comparison graphs and applies Graph Convolutional Networks (GCN) to learn node representations.
HGEN~\citep{Hgen} proposes a hierarchical heterogeneous graph encoder to model the heterogeneous relationships between number nodes and word nodes, while capturing long-range dependencies across different node types.
3) \textbf{Attention-based models} leverage the attention mechanism to identify key relationships between mathematical concepts. 
For instance, MATH-EN~\citep{MATH-EN} employs self-attention to capture long-range dependencies in math word problems, while Group-ATT~\citep{group-attn} uses multi-head attention to extract different types of MWP features.
4) By pre-training on a large text corpus, \textbf{pre-trained language models (PLMs)} acquire valuable world knowledge and develop strong language understanding capabilities, which are also advantageous for solving math word problems. For example, Generate \& Rank~\citep{generate-rank} introduces a novel ranking task for MWPs within a multi-task framework built on a generative PLM, effectively addressing the challenge of minor errors in mathematical expressions.

With the rise of LLMs, researchers are increasingly using them to solve MWPs. Compared to earlier deep learning approaches, LLMs offer several advantages: 1) improved language comprehension, including better handling of numerical values; 2) rich pre-trained knowledge, which includes a vast amount of mathematical knowledge; 3) powerful reasoning and text generation capabilities, allowing LLMs to generate natural language rationales, while traditional methods typically generate only mathematical expressions; and 4) emergent abilities in in-context learning and instruction-following, enabling LLMs to solve a wide range of math problems without needing specialized training on MWP datasets.

Research on using LLMs to solve MWPs can be categorized into two main approaches: fine-tuning and prompting. 1) Fine-tuning methods involve updating model parameters using annotated and synthetic data~\citep{distill_step_by_step,tailored_learning}. 2) Prompting-based approaches take advantage of the in-context learning and instruction-following abilities of LLMs, eliminating the need for model training. These methods enhance LLM reasoning by designing advanced prompts and agentic workflows~\citep{CoT,PoT,ToT}. Our work falls into the second category, which we will discuss in more detail in Section~\ref{reason_llm_prompt}.

\subsection{Reasoning with LLM Prompting}
\label{reason_llm_prompt}
It is widely recognized that complex reasoning problems are quite challenging for language models.
Such problems include mathematical reasoning~\citep{survey-dl-math,tora,RGFNet,NERHRT}, commonsense reasoning~\citep{CSQA}, multimodal reasoning~\citep{multimodal_reasoning,medical_reasoning}, symbolic reasoning~\citep{CoT} and logical reasoning~\citep{selection-inference}. 
The recently proposed Chain-of-Thought (CoT) prompting~\citep{CoT} enhances LLMs’ ability to handle complex reasoning by generating intermediate steps that lead to the final answer. 
Similarly, Zero-shot CoT (ZeroCoT)~\citep{ZeroCoT} generates reasoning steps using a simple prompt, ``Let's think step by step'', without requiring exemplars.
Program-aided language model (PAL)~\citep{PAL} and Program-of-Thought (PoT)~\citep{PoT} generate programs to represent the reasoning process and utilize a code interpreter to execute the programs.
These approaches have inspired various prompting techniques that further extend LLMs' reasoning capabilities.

Two main technical approaches have emerged from these developments:
1) The first type of methods adopt the idea of ``divide and conquer''. This type of methods decompose complex tasks into simpler subtasks.
For instance, Plan-and-Solve (PS) prompting~\citep{Plan-and-solve} devises a plan to divide the entire task into smaller subtasks, and then carry out the subtasks according to the plan. Least-to-Most (L2M)~\citep{L2M} and DecomP~\citep{Decomposed-Prompting} similarly decompose complex problems into simpler sub-problems, sequentially solving each to arrive at the final answer.
2) The second approach follows the ``try more'' principle, generating multiple potential solutions and selecting the most likely one.
Self-Consistency (SC)~\citep{SC} decoding strategy improves CoT by sampling multiple solutions in a single round and determining the final answer through majority voting. Progressive-Hint-Prompting (PHP)~\citep{PHP} iteratively solves problems across multiple rounds and utilizes previous answers as guidance for subsequent attempts.
Besides, Tree-of-Thought (ToT)~\citep{ToT} and SelfEval-Guided-Decoding~\citep{SelfEval-Guided-Decoding} sample multiple responses at each step and integrate step-wise self-evaluation to guide the generation of a whole solution. 

Despite their advancements, most existing approaches apply a fixed solver regardless of problem complexity, which may result in unnecessary computational overhead or sub-optimal performance.
Recent works have attempted to address this inefficiency. FrugalGPT~\citep{frugalgpt} and MoT-cascade~\citep{model_cascade} dynamically combine weaker and stronger models to reduce computational costs while maintaining performance.
Similarly, Adaptive Consistency (AC)~\citep{adaptive_consistency} dynamically adjusts the number of samples in SC~\citep{SC} based on a stopping criterion to minimize the sample budget.

However, these approaches focus on adjusting a single dimension, either LLM model or sample size. 
In contrast, our proposed framework offers a more comprehensive solution by adapting multiple aspects of the solver, including the LLM model, sample size, prompting technique, and decomposition granularity. 
This diversity enables the combination of different adaptation strategies to create various solver configurations, optimizing both cost-efficiency and performance.

\subsection{Automated Feedback for LLMs} 
Another relevant area of research is the generation of automated feedback for LLM responses. As categorized by~\citep{survey-auto-correct}, automated feedback can be derived from two main sources: self-feedback and external feedback. Self-feedback originates from the LLM itself, through techniques such as self-evaluation~\citep{self-refine,Self-Verification,RR}. External feedback, on the other hand, comes from external models~\citep{shepherd}, tools~\citep{critic}, evaluation metrics~\citep{maieutic}, or knowledge bases~\citep{Retrieval-Feedback}.

In our framework, the \textit{evaluation} module can incorporate various forms of automated feedback. For simplicity, our current implementation uses a self-consistency-based metric (i.e., \textit{consistency})~\citep{SC} to evaluate solution quality, focusing on the effectiveness of the \textit{adaptation} module.

\section{Preliminaries}

Let $\textbf{q} \in \mathcal{Q}$ and $\textbf{s} \in \mathcal{S}$ represent a reasoning problem and its corresponding solution, where $\mathcal{Q}$ is the space of problems and $\mathcal{S}$ is the space of solutions. The LLM is denoted as $f_\theta$, with $\theta$ representing the model weights, and the prompting method is denoted as $\textbf{p}$.

\paragraph{\textbf{Chain-of-Thought Prompting}.} In reasoning problem-solving, Chain-of-Thought (CoT) Prompting~\citep{CoT} enables LLMs to generate a solution $\textbf{s}$ (i.e., the reasoning process) and a final answer $\textbf{a}$, given a problem $\textbf{q}$, an LLM $f_\theta$, and a prompting method $\textbf{p}$. The final answer $\textbf{a}$ can be extracted from the reasoning process $\textbf{s}$. This process is formulated as: $\textbf{s}, \textbf{a} = f_\theta(\textbf{q}, \textbf{p})$, where $\textbf{s}$ consists of a sequence of reasoning steps $\textbf{s} = (\textbf{t}_1, \dots, \textbf{t}_i, \dots, \textbf{t}_k)$, with $\textbf{t}_i$ representing the $i$-th reasoning step (i.e., thought) and $k$ denoting the number of steps.

\paragraph{\textbf{Self-Consistency Strategy}.} The self-consistency strategy~\citep{SC} enhances CoT prompting by replacing greedy decoding with a sampling-based decoding method. It generates multiple reasoning paths and selects the most consistent answer by marginalizing over all paths. Let $s$ represent the number of reasoning paths (i.e., the sample size), a key factor influencing performance. The task can be reformulated as: $\{ \textbf{s}_1, \textbf{s}_2, \dots, \textbf{s}_s \}, \{ \textbf{a}_1, \textbf{a}_2, \dots, \textbf{a}_s \} = f_\theta(\textbf{q}, \textbf{p}, s)$, where the final answer is obtained by aggregating the sampled answers: $\textbf{a} = \text{Aggregate}({\textbf{a}_1, \textbf{a}_2, \dots, \textbf{a}_s})$, and \text{Aggregate ($\cdot$)} means choosing the most consistent one in SC.

\paragraph{\textbf{Decomposition-Based Prompting}.} In this type of approaches~\citep{L2M,Decomposed-Prompting}, the original problem is decomposed into multiple sub-questions, and the solution to the main problem is obtained by solving the sub-questions. This method consists of two main components: a decomposer and a sub-question solver. The problem decomposition can be formulated as: ${ \textbf{q}^1, \textbf{q}^2, \dots, \textbf{q}^m } = f^{decomp}_\theta(\textbf{q})$, where ${ \textbf{s}^1, \textbf{s}^2, \dots, \textbf{s}^d } = f^{solve}_\theta(\textbf{q}, \textbf{p}, { \textbf{q}^1, \textbf{q}^2, \dots, \textbf{q}^d })$. Typically, the answer $\textbf{a}$ can be extracted from the solution $\textbf{s}^d$ of the last sub-question. The number of sub-questions $d$ reflects the decomposition granularity; a higher $d$ indicates finer granularity. Decomposition granularity in our work is categorized into three levels: \textit{coarse}, \textit{medium}, and \textit{fine}.

\paragraph{\textbf{Solver}.} A solver refers to the combination of all elements involved in solving a problem, including the LLM model, prompting method, sample size, and decomposition granularity, etc. 
Formally, a solver is denoted as $\mathbf{A} = (\mathbf{m}, \mathbf{s}, \mathbf{p}, \mathbf{d})$, where $\mathbf{m}$ represents the model, $\mathbf{s}$ denotes the sample size, $\mathbf{p}$ specifies the prompting method, and $\mathbf{d}$ refers to the decomposition granularity. The decomposition granularity $\mathbf{d}$ can take one of the following values: \textit{coarse}, \textit{medium}, or \textit{fine}.
Our framework adapts one or more components of the solver dynamically to different types of questions, ensuring both efficiency and effectiveness.

\paragraph{\textbf{Pipeline}.} A pipeline is a sequence of solvers pre-selected based on performance evaluations on a validation set for each dataset. It is designed to maximize accuracy while minimizing computational costs. During inference, the adaptation module sequentially activates solvers from the pipeline as needed, dynamically adjusting the problem-solving strategy.
The pipeline is denoted as $\mathbf{L}$ and formulated as: \begin{equation*} \mathbf{L} = (\mathbf{A}_{1}, \mathbf{A}_{2}, \dots, \mathbf{A}_{i}, \dots, \mathbf{A}_{l}), \quad \mathbf{A}_{i} = (\textbf{m}_i, \textbf{s}_i, \textbf{p}_i, \textbf{d}_i) \end{equation*} where $\mathbf{A}_i$ represents the $i$-th solver in the pipeline, and $l$ denotes the maximum number of callable solvers. Each solver $\mathbf{A}_i$ is defined as a tuple consisting of several key elements: $\textbf{m}_i$ is the LLM model, $\textbf{s}_i$ is the sample size, $\textbf{p}_i$ is the prompting method, and $\textbf{d}_i$ is the decomposition granularity.

\section{The Adaptive-Solver Framework}
\label{method}

\begin{figure}[h]
\begin{center}
\includegraphics[width=\linewidth]{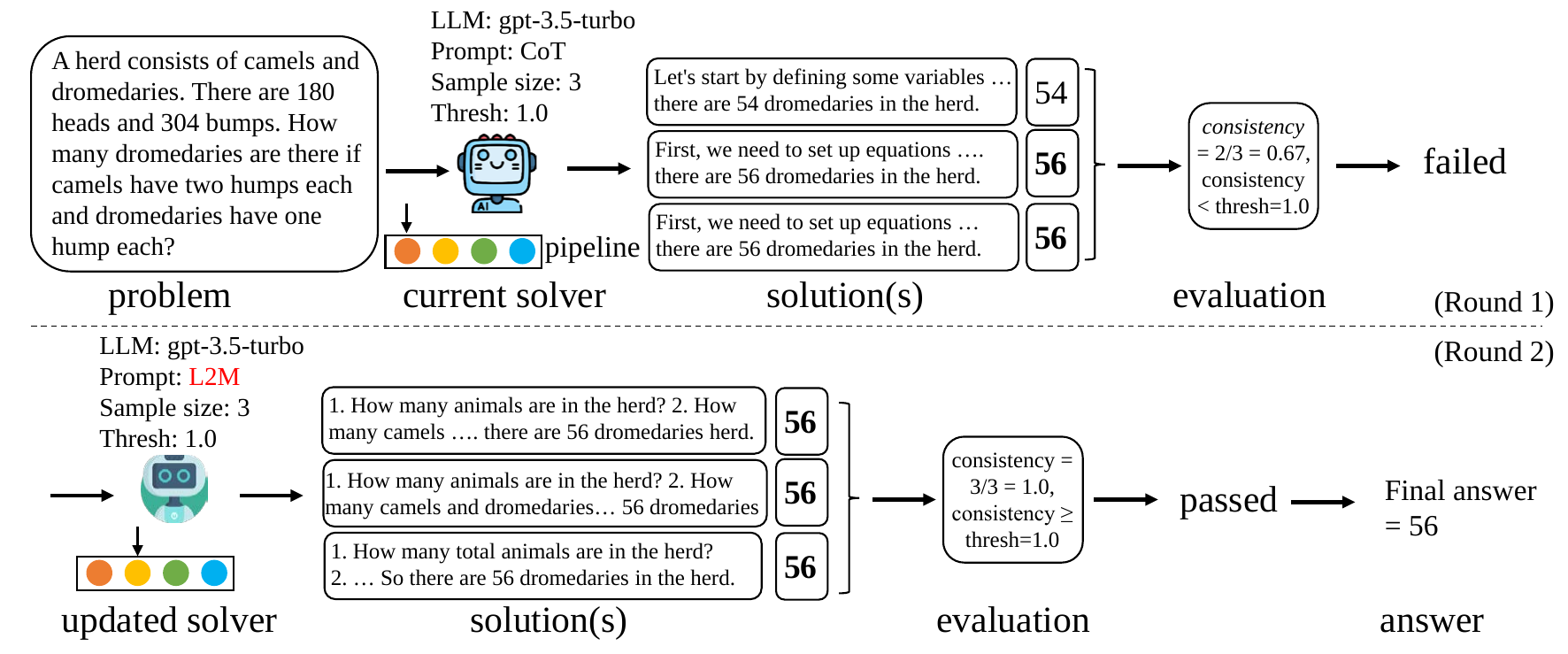}
\end{center}
\caption{
Overview of the Adaptive-Solver framework. It consists of two main modules: the \textit{evaluation} module assesses if the current solution meets the required criteria; if not, the \textit{adaptation} module adjusts the current solver by selecting the next solver from a predetermined pipeline. For simplicity, we illustrate a scenario with two solving rounds.
}
\label{framework}
\end{figure}

\textbf{Overview.} 
The Adaptive-Solver (AS) framework integrates multiple solvers and dynamically selects the most suitable one for each problem. It comprises two main modules: the \textit{evaluation} module and the \textit{adaptation} module. The overall workflow is depicted in Figure~\ref{motivation}(b), with an example case illustrated in Figure~\ref{framework}:

1) Given a problem, candidate solutions are generated by the current solver, and the \textit{evaluation} module checks whether the derived answer meets the specified criteria. If the criteria are satisfied or the maximum number of solving rounds is reached, the solving process terminates.

2) If the criteria are not met, the \textit{adaptation} module adjusts the solver and proceeds to the next round of solving, repeating step 1.
The \textit{adaptation} module activates solvers in sequence according to a predetermined pipeline.
Within this module, four key adaptation strategies are designed to provide guidance on how to adjust the solver, as explained in \cref{adaptation_strategies}.
An algorithm is proposed to automatically determine the optimal pipeline configuration using these strategies, as detailed in \cref{automatic_formulation}.

\subsection{Evaluation Module}
\label{evaluation component}
The \textit{evaluation} module determines whether the current solver is sufficient for the problem and decides when to trigger adaptation.
The study of self-consistency method \citep{SC} reveals a robust positive correlation between \textit{consistency} (measured by the proportion of the most frequent answer) and accuracy.  
This enables us to leverage \textit{consistency} to estimate the likelihood of the current answer being correct and reflect the confidence of model prediction.
Therefore, in our implementation of the proposed framework, each solver generates $s$ diverse solutions during a solving round, and the \textit{consistency} metric is computed. If the \textit{consistency} (i.e., number of the most frequent answer / $s$) reaches a predefined threshold $\theta$, the solving process terminates.
To maintain consistent rigor in evaluation, we set up distinct thresholds for different sample sizes.
\subsection{Adaptation Module}
\label{method: adaptation module}
The \textit{adaptation} module addresses the limitations of the ``one solver for all problems'' approach by dynamically adjusting the solver to suit different problems. This reduces computational costs and improves performance by identifying the most appropriate solver for each problem.

The \textit{adaptation} module operates in two phases: an optimization phase and an inference phase. During the optimization phase, the optimal solver pipeline is determined based on accuracy maximization over a validation set. This is achieved using an algorithm that integrates four adaptation strategies to automatically configure the pipeline. In the inference phase, the adaptation module sequentially activates solvers from the pipeline when adaptation is required. This two-phase approach ensures efficient and real-time solver adjustments. The four adaptation strategies and their integration into the solver pipeline are discussed below.
 
\subsubsection{Adaptation Strategies}
\label{adaptation_strategies}

\begin{figure}[h]
\begin{center}
\includegraphics[width=\linewidth]{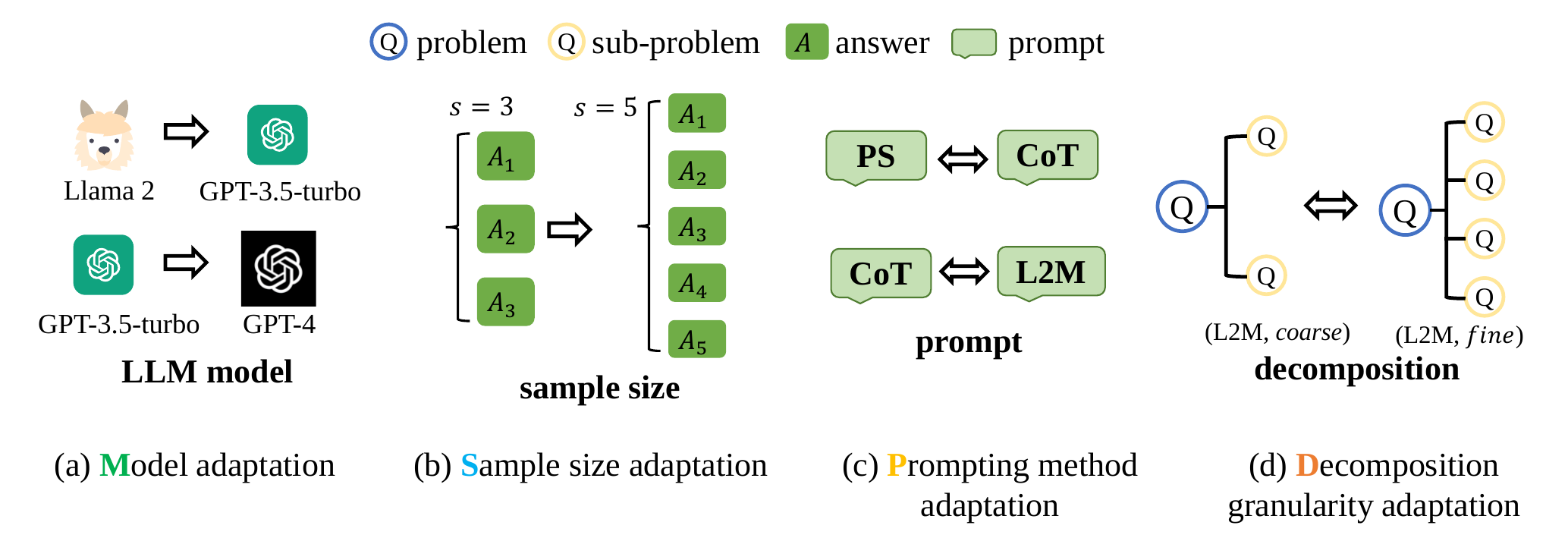}
\end{center}
\caption{Illustration of the four adaptation strategies. These strategies respectively consider the perspectives of the LLM model, sample size, prompt, and decomposition granularity.
}
\label{adaptations}
\end{figure}

1) \textbf{Model adaptation} (shown in Figure~\ref{adaptations}(a)) initializes the LLM model in a solver with a weaker yet cheaper LLM and gradually switches it to a more advanced yet more costly LLM when needed.
This adaptation strategy is designed to achieve high efficiency of simple problems and ensure the accuracy of solving complex problems.

2) \textbf{Sample Size Adaptation} (shown in Figure~\ref{adaptations}(b)) initiates the sample size within a solver with a small quantity, progressively augmenting it to improve the probability of accurately solving problems.

3) \textbf{Prompting Method Adaptation} (shown in Figure~\ref{adaptations}(c)) switches between various prompting methods in a solver to accommodate the unique characteristics of each problem.

4) \textbf{Decomposition Granularity Adaptation} (shown in Figure~\ref{adaptations}(d)) modulates the level of decomposition granularity of decomposition-based prompts utilized within a solver. This adaptation ensures optimal granularity for addressing problems of varying complexities. We design three variants of L2M~\citep{L2M} prompt, denoted as (L2M, \textit{coarse}), (L2M, \textit{medium}) and (L2M, \textit{fine}), each featuring different levels of decomposition granularity, ranging from coarse to fine. The only difference among them is the decomposition granularity in their demonstrations. These variant prompts can instruct an LLM to decompose the same problem at various levels of granularity.
Please refer to~\ref{Appendix:L2M variants construction} for more details about L2M's variants.

In this context, modifying decomposition granularity can be realized by adjusting prompting method, such as switching prompt (L2M, \textit{coarse}) to prompt (L2M, \textit{fine}). Therefore, the adaptation of decomposition granularity and prompting method are unified into a single process of adjusting prompting methods.

\subsubsection{Automatic Pipeline Configuration}
\label{automatic_formulation}

\begin{figure}[h]
\begin{center}
\includegraphics[width=\linewidth]{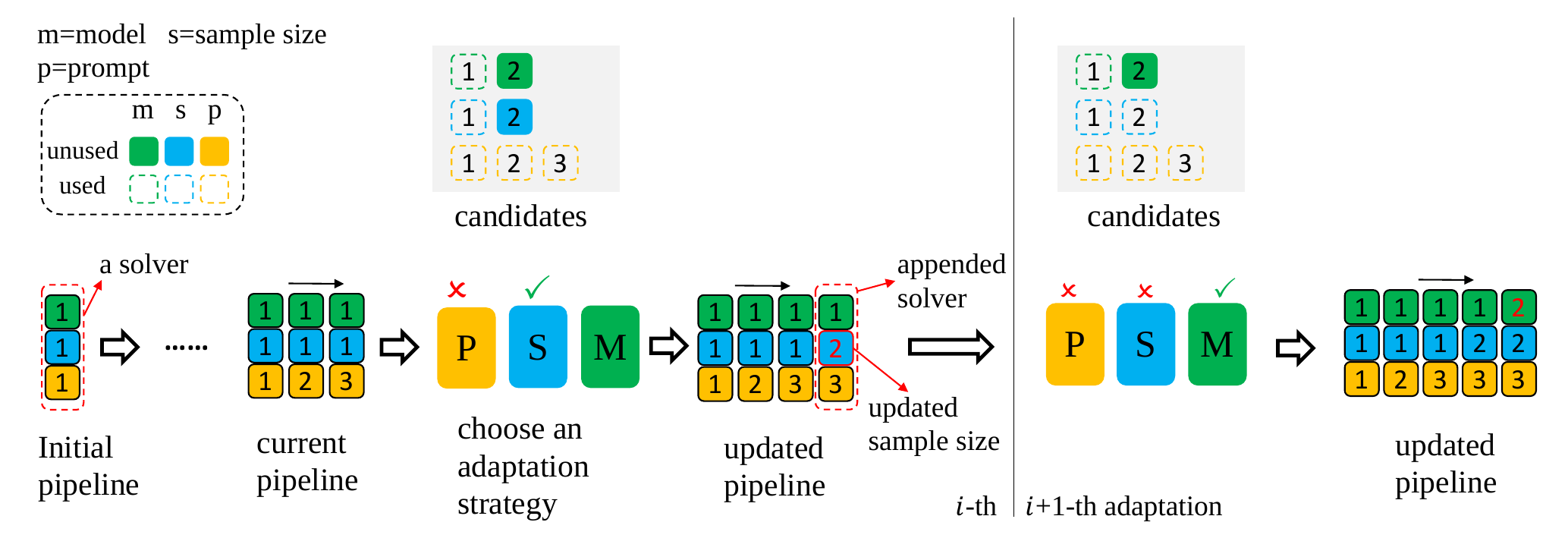}
\end{center}
\caption{Illustration of the process of pipeline configuration.
}
\label{pipeline}
\end{figure}

The goal of pipeline configuration is to efficiently identify the combination of solvers that maximizes performance while minimizing cost. Our analysis of the four adaptation strategies shows that performance improvements are typically accompanied by increased costs. For instance, \textit{Model Adaptation} (switching to a stronger LLM) can yield substantial performance gains but may increase costs by 20-30 times, while \textit{Prompting Method Adaptation} offers smaller performance gains with a cost increase of only 1-2 times.
To manage these trade-offs, we propose a heuristic algorithm that integrates the four adaptation strategies to ensure steady performance improvements with controlled cost increases. The algorithm adjusts the prompting method (or decomposition granularity), sample size, and LLM model within each solver in sequence.

The pipeline configuration process involves continual evaluation of the pipeline's performance and cost, which requires frequent interactions with the LLM’s API. This can lead to significant time and financial expenditure, making the configuration process unfeasible. To mitigate this, we initially gather and store multiple (e.g., 10) responses for every (LLM model, prompt) pair across all validation set problems. For subsequent evaluations, we sample responses from these stored records (denoted as $D_{val}$) for each solver, eliminating any further API calls.
This strategy renders the pipeline configuration process both time-efficient and cost-effective.

The optimal pipeline is gradually formed by sequentially appending solvers, as depicted in Figure~\ref{pipeline}. 
During each adaptation step, the last solver in the current pipeline is duplicated, and an adaptation strategy is chosen to update the solver by replacing either the LLM model, sample size, or prompting methods with corresponding options from $candidates$ (shown in Figure~\ref{pipeline}). If adding the new solver enhances performance, it is appended to the pipeline. 
The detailed configuration process is as follows:

1) \textbf{Initialize the pipeline}: Set the first solver by assigning the least expensive LLM model and sample size, and selecting the prompting method that achieves the highest accuracy on the validation set with the chosen model and sample size.

2) \textbf{Adjust the solver}: In each iteration, modify one aspect of the solver in the following order: prompting method (or decomposition granularity), sample size, and LLM model. Only one adjustment is made per iteration.  

   - If alternative prompting methods are available, select the most complementary method to the current one.  
   
   - If no suitable prompting method is found but alternative sample sizes are available, increase the sample size to the next level.  
   
   - If no sample size adjustment is possible, and alternative models are available, switch to the next stronger yet more expensive model. 

3) \textbf{Evaluate the new solver}: After each adaptation step, if the new solver improves performance, append it to the pipeline; otherwise, discard it.

4) \textbf{Repeat}: Continue steps 2 and 3 until no further adjustments are available, and treat the final pipeline configuration as the output.

For further technical details, please refer to Algorithms \ref{algorithm1}, \ref{algorithm2}, and \ref{algorithm3}.

\begin{algorithm}
\caption{Automatic Pipeline Configuration of the AS framework}
\label{algorithm1}
\begin{algorithmic}[1]
    \Require candidate models $M$, candidate sample sizes $S$, candidate prompts $P$, records $D_{val}$, map from sample size to threshold $map$
    \Require GetAccCost \Comment{Calculate accuracy and cost for a given solver (Algorithm\ref{algorithm2})}
    \Require AdjustSolver \Comment{Adjust the solver and update the pipeline (Algorithm\ref{algorithm3})}
    \Ensure $pipeline$
    \State $acc \gets 0$, $cost \gets 0$
    \State Initialize $pipeline$ with the cheapest LLM model, sample size and the prompt performing best on the validation set
    \While{True} \Comment{Append a new solver to $pipeline$ in each loop}
        \State $stop\_flag$, $pipeline$, $M$, $S$, $P$ = AdjustSolver($pipeline$, $wrong\_set$, $M$, $S$, $P$) \Comment{Adjust the current solver}
        \State $\_acc$, $\_cost$, $wrong\_set$=GetAccCost($pipeline$, $D_{val}$, $map$) \Comment{$wrong\_set$ is the set of questions that current $pipeline$ fails on}
        \If{$\_acc$ > $acc$}
           \State update $acc$ and $cost$ with $\_acc$ and $\_cost$
        \Else
            \State get rid of the last solver from $pipeline$
        \EndIf
        % \State $stop\_flag$, $pipeline$, $M$, $S$, $P$ = AdjustSolver($pipeline$, $wrong\_set$, $M$, $S$, $P$) \Comment{Adjust the current solver}
        \If{$stop\_flag$ is True} 
        \State break \Comment{This will break the loop}
        \EndIf
    \EndWhile
\end{algorithmic}
\end{algorithm}

\begin{algorithm}
\caption{GetAccCost}
\label{algorithm2}
\begin{algorithmic}[1]
    \Require $pipeline$, $D_{val}$, map from sample size to threshold $map$
    \Ensure $acc$, $total\_cost$, $wrong\_set$
    \State $\_cost \gets 0$, $num\_correct \gets 0$, $num\_problem \gets 0$
    \State $wrong\_set \gets \text{empty set}$
    \For{problem $q$ in $D_{val}$}
        \State $num\_problem \gets num\_problem + 1$
        \For{$solver$ in $pipeline$}
            % \State get model $m$, prompt $p$, sample size $s$ from $solver$
            % \State $answers$, $responses$ = SampleAnswers($q$, $m$, $p$, $s$)
            \State given $solver$ and $q$, sample corresponding $responses$ and $answers$ from $D_{val}$
            % \State $temp\_cost$ = CalculateCost($q$, $m$, $p$, $responses$)
            \State given $solver$ and $responses$, calculate cost $temp\_cost$
            \State $\_cost \gets \_cost + temp\_cost$
            % \State $answer$, $consisteny$=GetAnswerConsistency($answers$)
            \State given $answers$, find the most consistent $answer$ and calculate its $consistency$
            \If{$answer$ is evaluated to be correct}
                \State $num\_correct \gets num\_correct + 1$
            \Else
                \State add the current problem into $wrong\_set$
            \EndIf
            \State $\theta$ $\gets$ get the threshold corresponding to $s$ from $map$
            \If{$consistency \geq \theta$}
                \State break \Comment{stop activating subsequent solvers}
            \EndIf
        \EndFor
    \EndFor
    \State $\_acc$ = $num\_correct$ / $num\_problem$
\end{algorithmic}
\end{algorithm}

\begin{algorithm}
\caption{AdjustSolver}
\label{algorithm3}
\begin{algorithmic}[1]
    \Require $pipeline$, $wrong\_set$, candidate models $M$, candidate sample sizes $S$, candidate prompts $P$
    \Ensure $stop\_flag$, updated $pipeline$, $M$, $S$ and $P$
    \State copy the last solver of $pipeline$ to $solver$
    \State $stop\_flag \gets False$
    \If{$P$ is not empty}
        \State find the prompting $prompt$ performing the best on $wrong\_set$
        \State update the prompt in $solver$ with $prompt$
        \State get rid of $prompt$ from $P$
    \ElsIf{$S$ is not empty}
        \State increase sample size in $solver$ to the next level $sample\_size$
        \State get rid of $sample\_size$ from $S$
    \ElsIf{$M$ is not empty}
            \State upgrade the LLM model in $solver$ to the next stronger $model$
            \State get rid of $model$ from $M$
    \Else
        \State $stop\_flag \gets True$
    \EndIf
    \State append $solver$ to $pipeline$
\end{algorithmic}
\end{algorithm}

\subsubsection{Time Complexity of Pipeline Configuration Algortihm} \label{sec:complexity}

The complexity of adjusting the solver in each iteration is \(O(N_p + N_s + N_m)\), where \(N_p\), \(N_s\), and \(N_m\) represent the number of available prompts, sample sizes, and LLM models, respectively. Evaluating the updated pipeline after each adjustment requires \(O(n \times (N_p + N_s + N_m))\), where \(n\) is the number of validation problems. Therefore, the total time complexity of the algorithm, considering the number of iterations and the evaluation for each adjustment, is \(O(n \times (N_p + N_s + N_m)^2)\).

To improve efficiency, our pipeline configuration algorithm incorporates two key designs: (1) pre-saved responses for each solver, reducing redundant API calls, and (2) a structured adjustment process that avoids an unconstrained search over all \(N_p \times N_s \times N_m\) combinations, ensuring steady cost increments.

\subsection{Characteristic of the AS Framework}
This section highlights the key characteristics of the Adaptive-Solver (AS) framework, outlining its advantages and potential limitations.

\subsubsection{Advantages} 

\textbf{Multifaceted Adaptation}: Unlike existing approaches that uses a fixed solver or focus on adapting only a single element, such as sample size~\citep{adaptive_consistency} or LLM model~\citep{model_cascade}, the AS framework provides a holistic solution. 
It dynamically adjusts multiple dimensions—including the LLM model, sample size, prompting method, and decomposition granularity—enabling more efficient and accurate problem-solving across a wider range of tasks.

\textbf{Cost-Efficiency}: The AS framework cuts costs by using less resource-intensive models for simpler problems, achieving 46-85\% savings in inference costs compared to always relying on powerful models like GPT-4. This dynamic selection balances cost and performance effectively.

\textbf{Improved Performance}: By adapting the LLM model, sample size, prompting method, and decomposition granularity based on each task’s complexity, the AS framework outperforms static solvers by selecting the most suitable solver for different problems.

\subsubsection{Limitations}
\textbf{Inference Time}: Multi-round solving strategies used in our method may increase inference time, especially for complex problems. This can be alleviated by controlling the pipeline length to keep the time increase minimal.

\textbf{Validation Set Requirement}: An additional validation set is required to evaluate each solver’s performance and cost, and to pre-determine a pipeline of solvers for each dataset. This overhead can be reduced by using a small validation set, typically fewer than 200 samples, and selecting representative problems to enhance evaluation accuracy.

\section{Experiments}

In this section, we aim to answer the following questions via experiments. 
\textbf{Q1}: How effective is the AS framework in terms of \textit{accuracy} and \textit{cost} compared with baselines? 
\textbf{Q2}: How does each of the four adaptation strategies contribute to the performance?
\textbf{Q3}: How does the AS framework balance the cost and performance?
\textbf{Q4}: What are the benefits of integrating diverse solving strategies, and what are the underlying reasons?
\textbf{Q5}: Does the AS framework significantly increase inference time?

\subsection{Experimental Settings}
\subsubsection{Datasets} 
The proposed method is evaluated on 8 datasets covering three types of reasoning tasks. 1) Arithmetic Reasoning: GSM8K~\citep{gsm8k}, SVAMP~\citep{svamp}, AQuA~\citep{aqua}, AddSub~\citep{addsub}, SingleEq~\citep{singleeq} and MultiArith~\citep{multiarith}; 2) Commonsense Reasoning: CSQA~\citep{CSQA}; 3) Symbolic Reasoning: Last Letter Concatenation (LLC)~\citep{CoT}. 
Each dataset is split into a validation and a test set, detailed in Table~\ref{table:dataset}. The validation set facilitates to identify the optimal pipeline in our method, and the test set is for performance and cost comparison across methods.

\begin{table} 
\centering
\caption{Dataset statistics. CS: commonsense reasoning, Sym.: symbolic reasoning.}
\label{table:dataset}
\begin{tabular}{cccc}
    \hline
    Dataset & Domain & \# Validate & \# Test \\ 
    \hline
    GSM8K & Math & 200 & 1119 \\ 
    SVAMP & Math & 200 & 800 \\ 
    AQUA & Math & 50 & 204 \\
    AddSub & Math& 50 & 345 \\
    SingleEq & Math & 100 & 408 \\
    MultiArith & Math & 94 & 506 \\ 
    CSQA & CS & 200 & 1021 \\
    LLC & Sym. & 100 & 400 \\
    \hline
\end{tabular}
\end{table}

\begin{table} 
\centering 
\caption{Pipelines used in our AS-MSPD method, determined by the automatic pipeline configuration algorithm on the validation set. G3.5=gpt-3.5-turbo, G4=gpt-4. Z=ZeroCoT, P=PS, C=CoT, L=L2M, L1=(L2M, \textit{coarse}), L2=(L2M, \textit{medium}), L3=(L2M, \textit{fine}).} 
\begin{tabular}{c|c}
    \hline 
    Dataset & Pipeline\\
    \hline
    GSM8K & (G3.5, Z, 3), (G3.5, L1, 3), (G3.5, C, 3), (G3.5, Z, 5), (G3.5, Z, 10), (G4, Z, 1) \\
    SVAMP &  (G3.5, P, 3), (G3.5, L2, 3), (G3.5, Z, 3), (G3.5, L3, 5), (G3.5, P, 10), (G4, Z, 1) \\ 
    AQUA & (G3.5, P, 3), (G3.5, C, 3), (G3.5, Z, 3), (G3.5, L, 3) \\
    AddSub &  (G3.5, C), (G3.5, L3) \\
    SingleEq &  (G3.5, C, 3), (G3.5, L1, 3), (G3.5, P, 3) \\
    MultiArith & (G3.5, C, 3), (G3.5, L, 3), (G3.5, L3, 3) \\ 
    CSQA & (G3.5, Z, 3), (G4, Z, 1) \\
    LLC &  (G3.5, C, 3), (G3.5, L, 3), (G4, Z, 1)\\
    \hline
\end{tabular}
\label{table:pipeline}
\end{table}

\subsubsection{Baselines}
We compare the AS framework against baseline methods that use fixed solvers (i.e., consistent LLM model, sample size, prompting method, and decomposition granularity). These baselines include: (GPT4, $s$=1, ZeroCoT~\citep{ZeroCoT}) and all instances of ($m$, $s$, $p$), where $m\in$ \{GPT3.5\}, $s\in$ \{1, 3, 5, 10\}, $p\in$ \{ZeroCoT, PS~\citep{Plan-and-solve}, CoT~\citep{CoT}, L2M~\citep{L2M}\}. 
% When the LLM model is GPT4, the prompting method and sample size are only set to ZeroCoT and 1. This is because this setting is already 

\subsubsection{Evaluation Metrics} 
We evaluate all methods using \textit{answer accuracy} and \textit{API cost} on the test set. At the time of our experiments, GPT-3.5-turbo’s API cost was \$0.0015/1K prompt tokens and \$0.002/1K completion tokens, while GPT-4’s API cost was \$0.03/1K tokens and \$0.06/1K tokens.

\subsubsection{Implementation}
We set the temperature as 0 for the greedy decoding strategy and 0.7 for the methods with self-consistency strategy. We set different thresholds for different sample sizes in the evaluation module and the map from sample size to threshold is: \{1: 1.0, 3: 1.0, 5: 0.8, 10: 0.6\}.
Table~\ref{table:pipeline} presents the pipelines employed in our method (denoted as AS-MSPD) across different datasets.
All prompts used in this work are provided in Appendix~\ref{Appendix: prompts}.

\subsection{Main Results (Q1)}
\label{section: main results}

\begin{figure}[h]
\begin{center}
    \includegraphics[width=\linewidth]{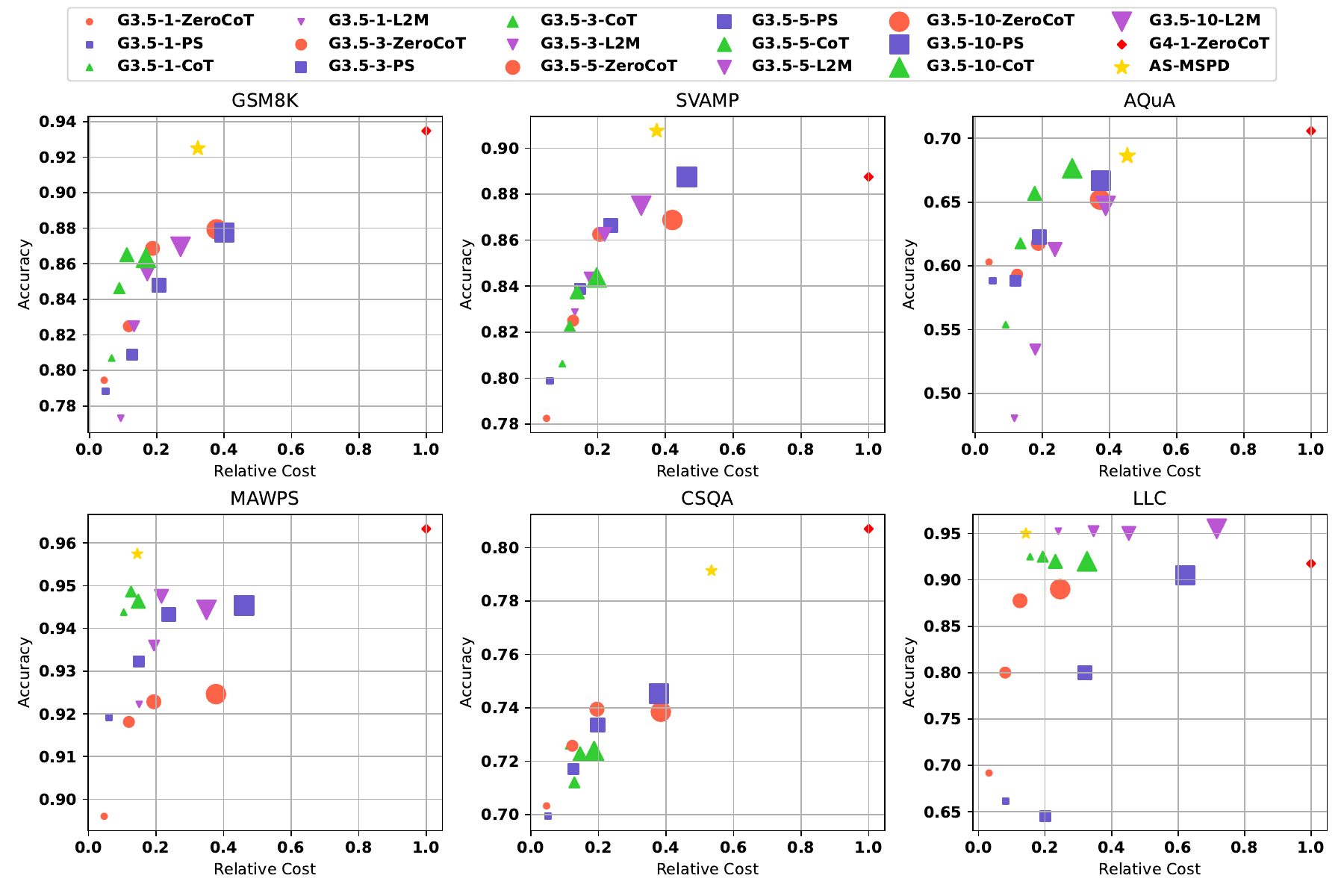}
\end{center}
\caption{Comparison of accuracy and cost across 8 reasoning datasets. MAWPS results are averaged over three datasets: Addsub, SingleEq and MultArith. 
Each point represents a method, with the same color and shape indicating the same model and prompt, and the point’s size reflects the (average) sample size.
G3.5: GPT-3.5-turbo-0301. G4: GPT-4-0613.
Relative cost represents the API cost compared with that of G4-1-ZeroCoT. 
}  
\label{fig_main_results}
\end{figure}

Figure~\ref{fig_main_results} illustrates the comparison of various methods in terms of \textit{accuracy} and \textit{cost}. We can observe that:

\textbf{AS-MSPD significantly reduces cost while maintaining performance comparable to GPT4}.
When comparing GPT4-1-ZeroCoT with GPT3.5-3-ZeroCoT, the former significantly outperforms the latter, leading by approximately 6-16\%. Nonetheless, this enhanced performance comes at a substantially higher cost, about 7-13 times more expensive. By contrast, AS-MSPD matches or slightly exceeds GPT4’s performance while substantially reducing API costs by about 46-85\% compared to using GPT4 alone.

\textbf{AS-MSPD outperforms all other methods within the same cost range}.
Across all datasets, AS-MSPD consistently outperforms baseline methods with equivalent budgets.  
For instance, on the dataset GSM8K, AS-MSPD (92.49\%) outperforms the best baseline G3.5-10-ZeroCoT (87.93\%) by 4.56\%, and on the dataset SVAMP, AS-MSPD (90.75\%) surpasses G3.5-10-PS (88.75\%) by 2\%. This underscores our method’s effectiveness in enhancing the reasoning ability of LLMs through the dynamic selection of the most appropriate solving strategies.

\textbf{Increasing sample size alone does not guarantee cost-effective performance improvement}.
The self-consistency strategy produces multiple solutions within a single solving round and selects the most consistent answer. 
While accuracy generally improves with larger sample sizes, the benefit tends to plateau, and in some cases, may even degrade performance (e.g., GPT3.5-CoT and GPT3.5-L2M in MAWPS).
This suggests that simply increasing the sample size is insufficient for further performance enhancement, and adjustments in the LLM model and prompting method should also be considered.

\textbf{Weaker models can complement stronger models by enhancing overall performance}. On datasets like SVAMP and LLC, our approach, which integrates GPT3.5 and GPT4, surpasses the performance of GPT4 alone. This indicates that GPT3.5, when equipped with appropriate sample size and prompting method, can solve problems that GPT4 cannot, and at a significantly lower cost. This underscores the potential of leveraging multiple LLM models to reduce costs while maintaining or even enhancing performance.

\subsection{Ablation Study of the Adaptation Strategies (Q2)}

\begin{figure}[h]
\centering
\includegraphics[width=\linewidth]{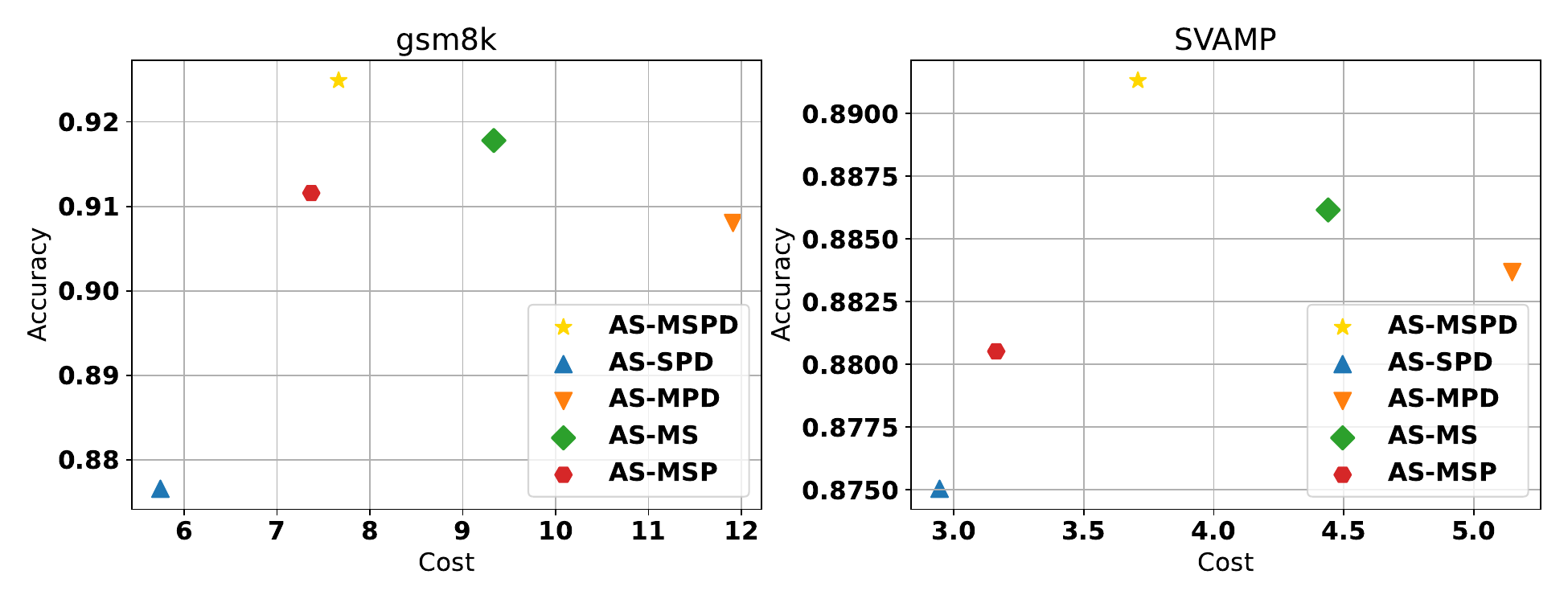}
\caption{Ablation study on different adaptation strategies.} 
\label{figure_ablation}
\end{figure}

In the adaptation module of our AS framework, we design four adaptation strategies. 
To evaluate the contribution of each adaptation strategy in the AS framework, we designed four variants of the framework, each omitting one adaptation strategy. 
These variants are compared to assess their impact on performance. The four variants are:

\begin{itemize}
    \item \textbf{AS-SPD}: Omits \textit{model adaptation}, consistently using the GPT-3.5 model.
    \item \textbf{AS-MPD}: Disregards \textit{sample size adaptation}, fixing the sample size at 10.
    \item \textbf{AS-MS}: Excludes both \textit{prompting method adaptation} and \textit{decomposition granularity adaptation}, using only the ZeroCoT prompt.
    \item \textbf{AS-MSP}: Ignores \textit{decomposition granularity adaptation}.
\end{itemize}

Figure~\ref{figure_ablation} shows the results of this ablation study. From the analysis, we observe the following:

\textbf{Each adaptation strategy contributes to performance increase or cost reduction.}
1) AS-MSPD achieves the best balance between performance and cost. This highlights the advantage of integrating all four adaptation strategies, which allows the framework to optimize both accuracy and cost-efficiency.
2) AS-SPD performs the worst, as it lacks \textit{model adaptation} and relies solely on the less powerful GPT-3.5 model. This highlights the significant role of \textit{model adaptation} in modulating overall performance.
3) AS-MPD incurs the highest cost, as it fixes the sample size at 10 without leveraging smaller, less expensive sample sizes. This demonstrates the value of \textit{sample size adaptation} in cost management.
4) AS-MS incurs higher expenses than AS-MSPD because it directly changes the LLM model and sample size, quickly driving up costs. In contrast, AS-MSPD begins with adapting prompts, resulting in a more gradual increase in costs.
5) AS-MSP does not perform as well as AS-MSPD, indicating the effectiveness of \textit{decomposition granularity adaptation} in further enhancing performance.

\subsection{Balance between Performance and Cost (Q3)}
\begin{figure}[ht]
\begin{center}
    \subfigure[GSM8K]{
        \includegraphics[width=0.475\linewidth]{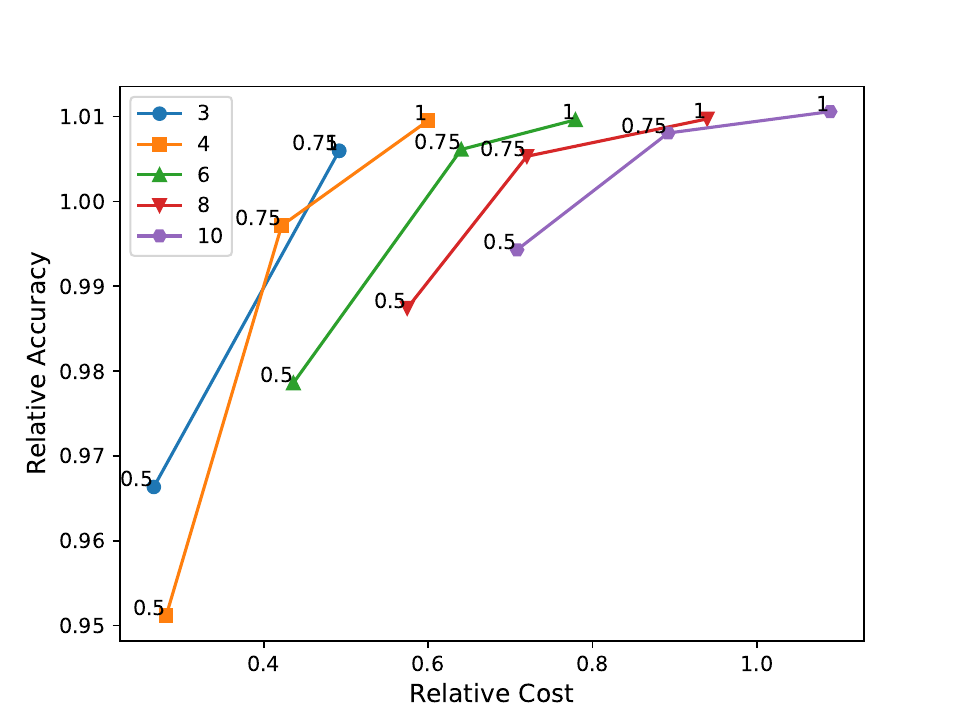}
        \label{figure_balance_gsm8k}
    }
    \subfigure[SVAMP]{
        \includegraphics[width=0.475\linewidth]{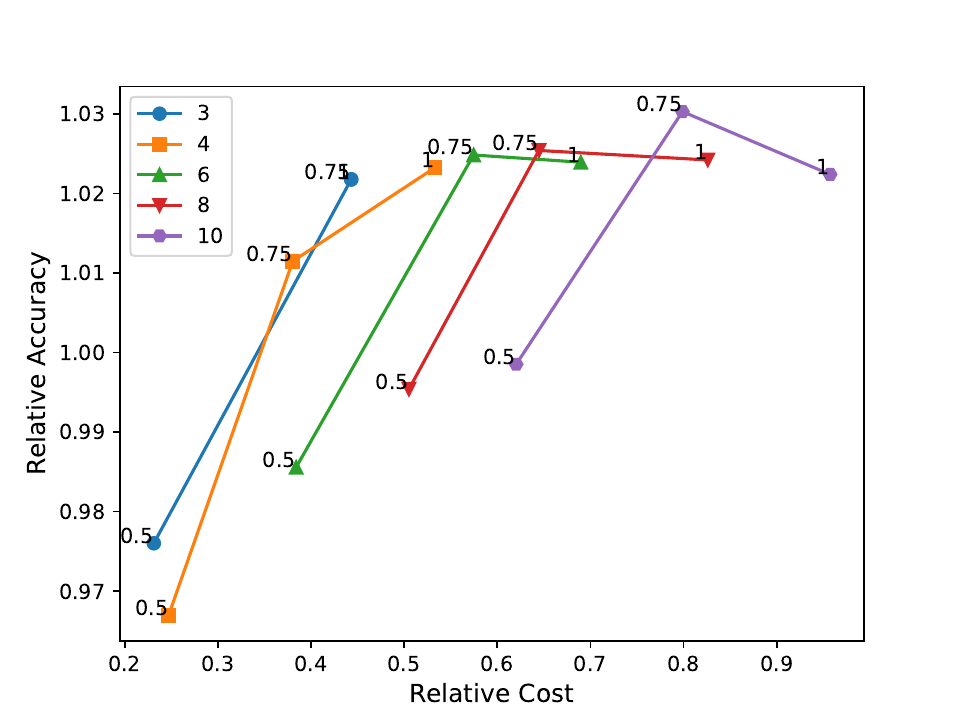}
        \label{figure_balance_svamp}
    }
\end{center}
    \caption{Variations in performance and cost across different sample sizes and thresholds, on the GSM8K and SVAMP dataset. Relative cost (accuracy) represents the API cost (accuracy) compared with that of G4-1-ZeroCoT.}
    \label{figure_balance}
\end{figure}

The evaluation module in the AS framework employs a consistency-based method, controlled by two key hyper-parameters: sample size $s$ and threshold $\theta$. We investigate how variations in these parameters influence both performance and cost.
To streamline our discussion, we focus exclusively on \textit{model adaptation} and set the pipeline as [(GPT3.5, $s$, ZeroCoT), (GPT4, 1, ZeroCoT)], with $s$ ranging among $\{3, 4, 6, 8, 10\}$ and $\theta$ among $\{0.5, 0.75, 1\}$.
From figure~\ref{figure_balance}, we find that:

\textbf{Our method has flexibility to balance performance and cost by tuning the hyper-parameters.}
1) Increasing the threshold for the same sample size (indicated by the same color) improves performance but also raises costs. This is because a higher threshold enforces stricter evaluation, causing more problems to be passed to a stronger but more expensive model for resolution. 
2) With the same threshold, when the threshold is at 0.5, bigger sample sizes yield higher performance. However, with the threshold at 0.75, a sample size of 3 can achieve almost the same performance as a sample size of 10.
This suggests that beyond a certain threshold, enlarging the sample size does not markedly improve performance but still raises costs significantly.
These observations highlight a trade-off between performance and cost. 
By tuning the hyper-parameters, our method provides the flexibility to adjust this balance to suit practical needs, offering tailored solutions for varying budget and accuracy requirements.

\subsection{Effect of Integrating Diverse Methods (Q4)}
\begin{figure}[ht]
\begin{center}
    \includegraphics[width=\linewidth]{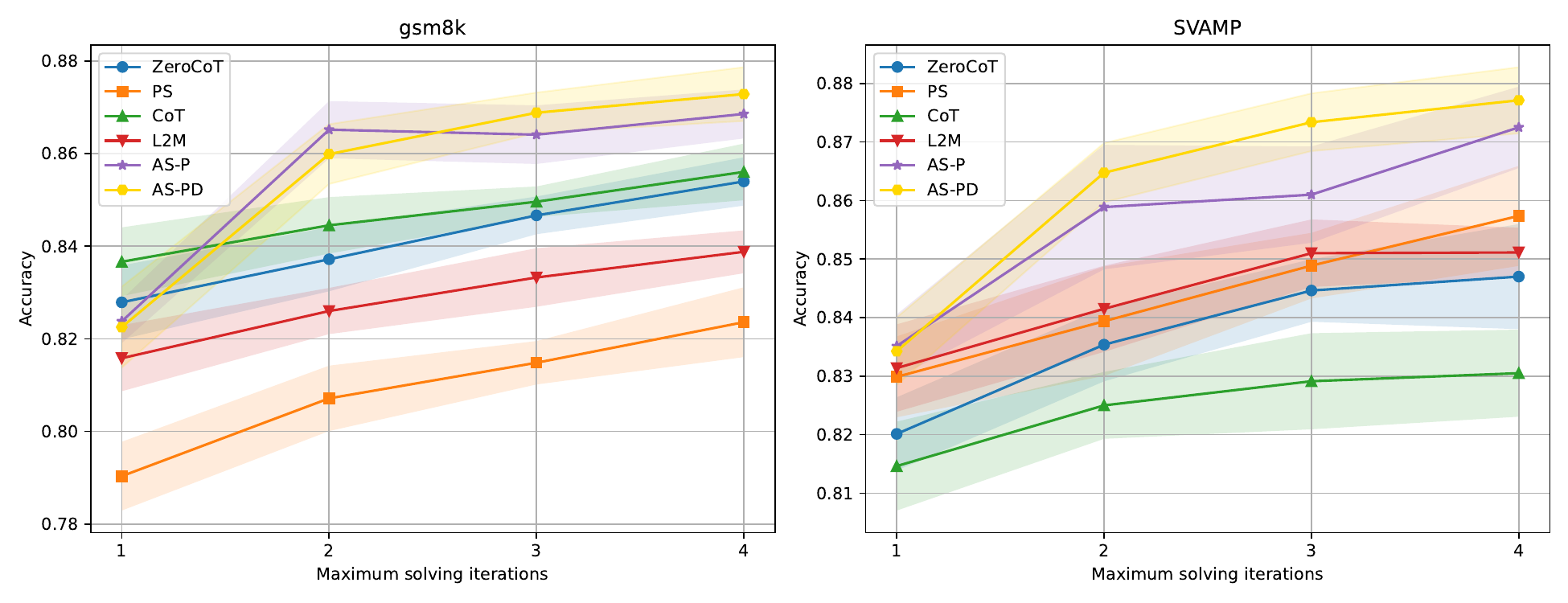}
\end{center}
\caption{Accuracy changes across different maximum solving iterations.} 
\label{figure_multi_round}
\end{figure}

To further explore the efficacy of “adaptation” (i.e., the adjustment of solving strategies), we compare our adaptive method against non-adaptive baselines that maintain a static solving strategy. We utilize two variants of the AS framework, AS-P and AS-PD, as the adaptive methods. All methods use GPT-3.5 as the LLM model and 3 as the sample size.
To eliminate the impact of multi-round solving strategy, we allow all methods to solve problems for multiple rounds. We observe accuracy fluctuations across different max number of solving rounds. 
The results are depicted in Figure~\ref{figure_multi_round}.

\textbf{Multi-round solving strategy benefits to the improved performance.} Both non-adaptive baselines and adaptive methods show performance improvements as the max number of solving rounds increases. This suggests that the \textit{evaluation} module and multi-round solving strategy alone in our AS framework can effectively boost performance.

\textbf{The adaptive methods exhibit superior and more consistent performance compared to non-adaptive baselines.} Both AS-P and AS-PD outperform non-adaptive baselines. 
Specifically, AS-PD leads the most effective non-adaptive baseline by approximately 2\% on both the GSM8K and SVAMP datasets.
Besides, the optimal non-adaptive method varies depending on the dataset. For instance, CoT and ZeroCoT outperform L2M and PS on the GSM8K dataset but performance worse on the SVAMP dataset. In contrast, adaptive methods consistently perform well across diverse datasets.
These observations indicate that our method enhances performance by dynamically adapting the solver, rather than simply by increasing solving iterations. Furthermore, its performance exhibits greater stability across diverse datasets.

% \subsection{Complementary effects among diverse solving strategies}
\subsection{Why Integrating Diverse Methods Benefits (Q4)}
We further explore why integrating multiple solving methods improves performance.
We focus on two adaptation strategies, \textit{prompting method adaptation} and \textit{decomposition granularity adaptation}, to conduct this analysis.

\textbf{Prompting method adaptation combines the advantages of different prompting methods.} 
Our analysis of \textit{prompting method adaptation} utilizes a specified pipeline [CoT, L2M], with LLM model as GPT3.5 and sample size as 3. We analyze the relation between the questions answered correctly by [CoT, L2M] and those questions correctly answered by CoT and L2M respectively. 
As detailed in Table~\ref{table-analysis-prompt-adaptation}, we categorize all problems into four distinct groups, and calculate [CoT, L2M]'s accuracy and the utilization of CoT and L2M in [CoT, L2M] for each group.
We find that [CoT, L2M] consistently delivers correct answers when both CoT and L2M succeed. 
Notably, in situations where either CoT or L2M succeeds, [CoT, L2M] addressed the majority (60\%-70\%) of those problems.
These findings indicate that \textit{prompting method adaptation} merges the advantages of both prompting methods, leading to a performance boost.

\begin{table}[ht]
\caption{Analysis of \textit{prompting method adaptation}. CoT \checkmark, L2M \ding{55}: CoT succeeds and L2M fails. \# problem: number of problems in each group.
\# correct: number of problems that [CoT, L2M] succeeds on.
}
\label{table-analysis-prompt-adaptation}
\begin{center}
    \begin{tabular}{ccccccc}
    \hline
        Dataset & CoT & L2M & \# problem & \# correct & \makecell{\# CoT \\ usage} & \makecell{\# L2M \\ usage} \\ \hline
        \multirow{4}*{GSM8K} & \checkmark & \checkmark & 995 & 984 & 884 & 11\\ 
        & \checkmark & \ding{55} & 123 & 76 & 62 & 61 \\
        & \ding{55} & \checkmark & 84 & 56 & 31 & 53 \\
        & \ding{55} & \ding{55} & 117 & 25 & 50 & 67\\
        \hline
        \multirow{4}*{SVAMP} & \checkmark & \checkmark & 762 & 755 & 692 & 70\\
        & \checkmark & \ding{55} & 51 & 34 & 31 & 20\\
        & \ding{55} & \checkmark & 94 & 65 & 34 & 60 \\
        & \ding{55} & \ding{55} & 93 & 15 & 45 & 48\\
    \hline
    \end{tabular}
\end{center}
\end{table}

\begin{figure}[h]
\centering
    \includegraphics[width=0.65\linewidth]{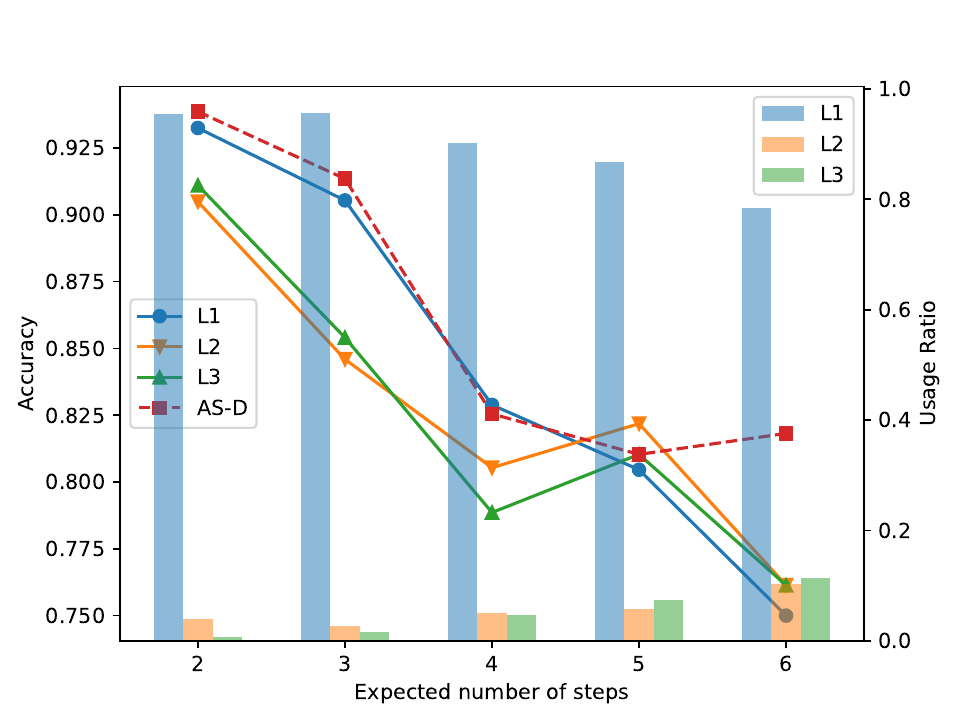}
    \caption{Analysis of \textit{decomposition granularity adaptation} on GSM8K. 
    }
    \label{figure_decomposition_analyze}
\end{figure}

\textbf{Decomposition granularity adaptation tailors decomposition granularity to problems with varied difficulties.}
Our analysis of \textit{decomposition granularity adaptation} utilizes a specified pipeline [L1, L2, L3] (denoted as AS-D), where $\rm{L_1}$ = (L2M, \textit{coarse}), $\rm{L_2}$ = (L2M, \textit{medium}), $\rm{L_3}$ = (L2M, \textit{fine}).
The LLM model and sample size are set to GPT3.5 and 3. 
We analyze how \textit{decomposition granularity adaptation} selects appropriate decomposition granularity for problems of varying difficulty, as illustrated in Figure~\ref{figure_decomposition_analyze}.
The problem difficulty is measured by the number of expected solving steps, provided by the GSM8K dataset.
The line chart segment reveals that for problems necessitating fewer than five steps, a coarse-grained decomposition L1 outperforms finer-grained decomposition L2, L3. Conversely, as problem difficulty increases, L2, L3 demonstrate superior performance than L1. This suggests that problems of varying difficulty require different levels of decomposition.
Furthermore, the bar chart segment highlights a progressive increase in the employment of finer-grained L2, L3 in AS-D in response to heightened problem difficulty.
These observations affirm that \textit{decomposition granularity adaptation}, by selectively employing decomposition strategies of varying granularity, can enhance LLM's performance.

%%%%%%%%%%%%%%%%%%%%%%%%%%%%%%%%%%%%%%%%%%%%%%%%%%%%%%%%%%%%%%%%%%%%%%%%

\subsection{Analysis of Time Efficiency (Q5)}
\label{main:efficienct analysis}
\textbf{Inference time of AS-MSPD does not increase significantly.}
Table~\ref{table: time efficiency} presents the average inference time per problem of various methods and the average solving rounds of our method AS-MSPD.
Despite employing a multi-round solving strategy, AS-MSPD shows no significant increase in average inference time. Specifically, the average inference time of AS-MSPD is approximately 1.45 times that of G4-Z-1 and even lower than that of G3.5-Z-10. The average solving round of AS-MSPD is 1.68, indicating that our method typically requires interaction with the LLM API fewer than 2 times on average to solve a problem. This is because the majority of problems can be resolved by the initial solver, which is cheaper and faster, with subsequent solvers being invoked only in a few necessary cases.

\begin{table}[h]
\centering
\caption{Inference time (seconds) comparison. G3.5: GPT3.5. G4: GPT4. Z: ZeroCoT. \{1, 3, 5, 10\} are sample sizes. Relative time: time of AS-MSPD / time of G4-Z-1. \# Average call: average solving rounds of AS-MSPD.}
\label{table: time efficiency}
    \begin{tabular}{ccccccc}
    \hline
    Method & GSM8K & AQuA & CSQA & LLC & Average \\
    \hline
    G3.5-1-Z & 4.6 & 4.2 & 3 & 1.8 & 3.4 \\
    G3.5-3-Z & 7.5 & 5.7 & 4.7 & 3.1& 5.58\\
    G3.5-5-Z & 10.8 & 8.3 & 6.2 & 4.1 & 7.8 \\
    G3.5-10-Z & 17.6 & 9.9 & 8.8 & 5.1 & 11.58 \\
    G4-1-Z & 9.3 & 8.6 & 5.6 & 6.1 & 7.04 \\
    \hline
    AS-MSPD & 15.3 & 13.8 & 7.1 & 2.1 & 10.26 \\
    % \hline
    Relative time & 1.65 & 1.60 & 1.27 & 0.34 & 1.45 \\
    \# Average call & 1.81 & 2.46 & 1.38 & 1.06 & 1.68 \\
\hline
    \end{tabular}
\end{table}

%%%%%%%%%%%%%%%%%%%%%%%%%%%%%%%%%%%%%%%%%%%%%%%%%%%%%%%%%%%%%%%%%%%%%%%%

\section{Conclusion}
We introduce the Adaptive-Solver (AS) framework, designed to dynamically adapt solving strategies for LLMs across diverse reasoning scenarios, allowing for flexible allocation of test-time computational resources.
Central to this framework are two modules: the initial \textit{evaluation} module, which assesses the reliability of a given solution, and the subsequent \textit{adaptation} module, which adjusts the solver if the reliability evaluation fails.
Herein, four adaptation strategies are leveraged together to achieve multi-faceted adaptations. 
Additionally, we designed an efficient pipeline configuration algorithm that automatically determines the optimal solver pipeline, enabling real-time adjustments of solver in the \textit{adaptation} module.
Our experimental results underscore the framework's effectiveness. Specifically, AS-MSPD significantly reduces API costs by up to 85\%, retains performance comparable to GPT4, and surpasses all cost-comparable baselines.
This framework propels us into a promising direction in dynamic strategy selection for LLMs. 
Viewing from a higher point, every solver – be it model, prompting, decomposition, or augmented tools – can be regarded as a potential candidate in the component pool. The LLMs, armed with this framework, exhibit the flexibility to dynamically compose selected candidates, paving the way to optimal solution paths. 

%%%%%%%%%%%%%%%%%%%%%%%%%%%%%%%%%%%%%%%%%%%%%%%%%%%%%%%%%%%%%%%%%%%%%%%%

\section*{Acknowledgments}
This work is supported by the National Natural Science Foundation of China (62472461), and the Guangdong Basic and Applied Basic Research Foundation (2022A1515011690).

\appendix
\section{Appendix}
% Appendix sections are coded under \verb+\appendix+.

% \subsection{Approach for constructing the prompt of L2M's variants in \textit{decomposition granularity adaptation}}
\subsection{Introduction of L2M's variant prompts used in decomposition granularity adaptation}
\label{Appendix:L2M variants construction}

% \subsection{L2M's variants for decomposition granularity adaptation}
% \label{Appendix_prompt_l2m_variants}
\textbf{The three L2M's variant prompts mainly differ from the decomposition granularity.}
For example, facing the same problem, the prompt (L2M, $coarse$) may break it down into 2-3 sub-questions, the prompt (L2M, $meidum$) may decompose it into 4-5 sub-questions, and the prompt (L2M, $fine$) may decompose it into 6-8 sub-questions. See specific prompts of them in Figure~\ref{prompt_l2m_coarse_math}, Figure~\ref{prompt_l2m_medium_math} and Figure~\ref{prompt_l2m_fine_math}.
In addition, the difference between them and L2M lies in: L2M lacks precise control over decomposition granularity in its demonstrations, leading to a blend of various granularities. Conversely, in the demonstrations of these variants, the decomposition granularity is either coarse, medium, or fine, depending on the specific variant.

\begin{figure}[ht]
\centering
\includegraphics[width=\linewidth]{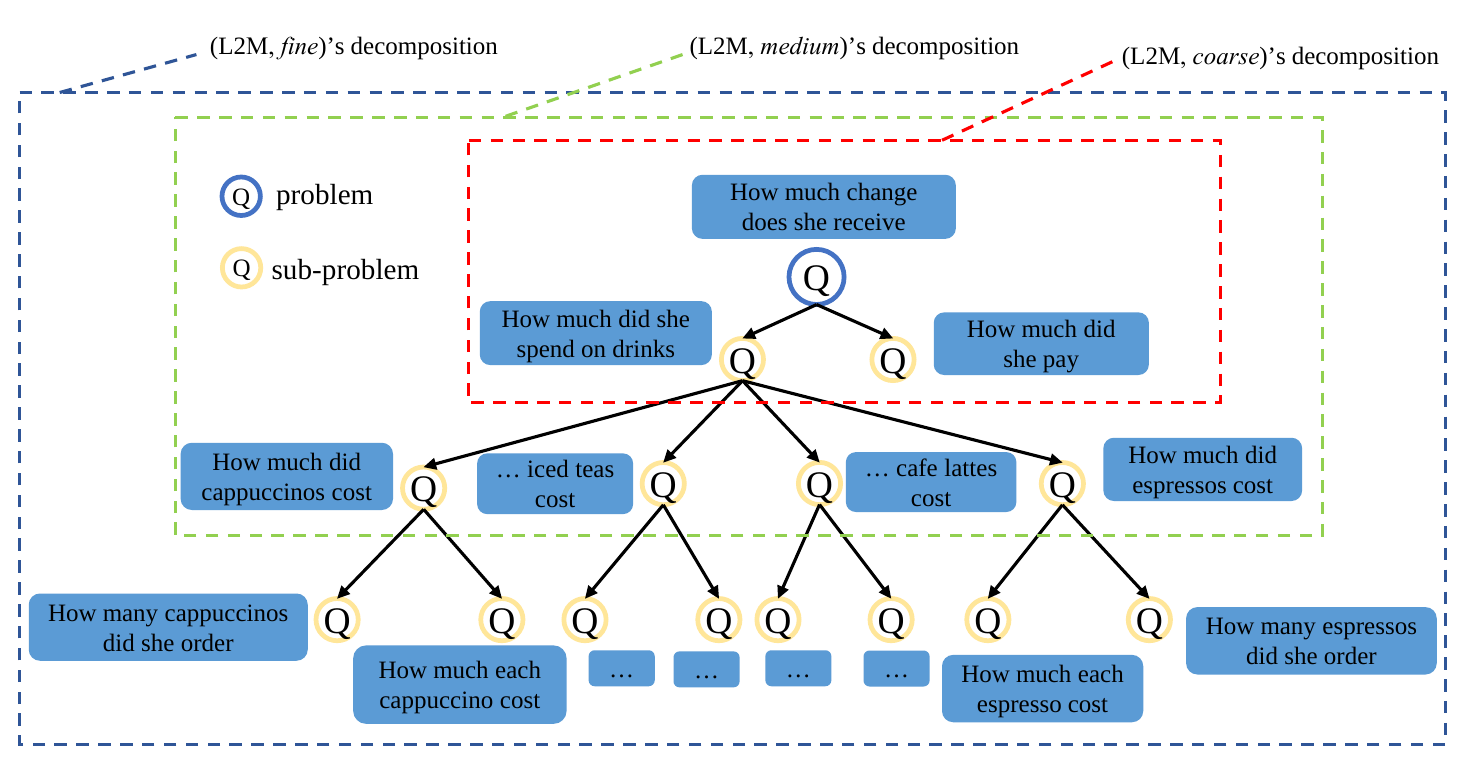}
\caption{Illustration of hierarchical decomposition.} 
\label{figure_prompt_construct}
\end{figure}

\textbf{Approach for constructing the prompt of L2M's variants.} To illustrate the construction process, consider the following example question: \textit{Cappuccinos cost \$2, iced teas cost \$3, cafe lattes cost \$1.5 and espressos cost \$1 each. Sandy orders some drinks for herself and some friends. She orders three cappuccinos, two iced teas, two cafe lattes, and two espressos. How much change does she receive back for a twenty-dollar bill?}

L2M does not control the decomposition granularity deliberately and its decomposition for the example question is as follows: \textit{1. How much did the cappuccinos cost in total? 2. How much did the iced teas cost in total? 3. How much did the cafe lattes cost in total? 4. How much did the espressos cost in total? 5. How much did Sandy spend on drinks? 6. How much change does she receive back for a twenty-dollar bill?}

To construct L2M’s variants, we \textbf{first decompose the question hierarchically, as shown in Figure~\ref{figure_prompt_construct}}.

1) First, we extract the problem and sub-problems from the first layer of decomposition. Then, serialize them from bottom to top to obtain the sequence of sub-problems in (L2M, \textit{coarse})’s prompt: \textit{1. How much did Sandy spend on drinks? 2. How much change does she receive back for a twenty-dollar bill?}

2) Similarly, we extract the problem and sub-problems from the first two layers of decomposition and then serialize them to obtain the sequence of sub-problems in (L2M, \textit{medium})’s prompt: \textit{1. How much did the cappuccinos cost in total? 2. How much did the iced teas cost in total? 3. How much did the cafe lattes cost in total? 4. How much did the espressos cost in total? 5. How much did Sandy spend on drinks? 6. How much change does she receive back for a twenty-dollar bill?}

3) Likewise, we extract the problem and sub-problems from the three layers of decomposition and serialize them to obtain the sequence of sub-problems in (L2M, \textit{fine})’s prompt: \textit{1. How many cappuccinos did Sandy order? 2. How much did the cappuccinos cost in total? 3. How many iced teas did Sandy order? 4. How much did the iced teas cost in total? 5. How many cafe lattes did Sandy order? 6. How much did the cafe lattes cost in total? 7. How many espressos did Sandy order? 8. How much did the espressos cost in total? 9. How much did Sandy spend on all drinks in total? 10. How much change does she receive back for a twenty-dollar bill?}

\textbf{Our method of constructing L2M's variants can indeed control the granularity in decomposition.}
Table~\ref{table: average number of subproblems} demonstrates the average number of sub-problems obtained by using L2M and L2M's variants.
We observe that finer-grained decomposition prompt indeed leads to a greater number of subproblems on average on the same dataset. 
This validates the effectiveness of controlling the granularity in the actual problem decomposition by modulating the granularity in the exemplars.
\begin{table*}[h]
\centering
\renewcommand\arraystretch{1.25}
\caption{Average number of sub-problems of various decomposition prompting methods.}
\label{table: average number of subproblems}
    \begin{tabular}{cccccccc}
    \hline
    Method & GSM8K & SVAMP & MultiArith & AddSub & SingleEq & AQuA & Average\\
    \hline
    L2M & 3.61 & 2.76 & 2.80 & 2.51 & 2.63 & 3.08 & 2.90\\
    (L2M, \textit{coarse}) & 2.60 & 1.88 & 2.06 & 1.73 & 1.77 & 2.19 & 2.04\\
    (L2M, \textit{medium}) & 3.6 & 2.76 & 2.73 & 2.44 & 2.54 & 2.74 & 2.80\\
    (L2M, \textit{fine}) & 4.46 & 3.56 & 3.51 & 2.85 & 3.15 & 3.57 & 3.52\\
    \hline
    \end{tabular}
\end{table*}

\subsection{Full sets of Prompts}
\label{Appendix: prompts}
We below provide all the prompts used in this work. For each prompt, if the response does not contain the phrase ``answer is'', we concatenate the question, response, and the phrase ``Therefore, the answer is'' before calling the API again to generate a concise response with the answer. For brevity and space considerations, only one example per prompt is shown below.
% We apply the same 4 examples for all few-shot prompting methods, although different prompting methods may necessitate varying modifications based on these exemplars. 

% \subsubsection{Zero-shot-CoT (ZeroCoT)}

\begin{figure}[h]
\begin{tcolorbox}
\small

% \textbf{Zero-shot-CoT (ZeroCoT): Prompt for all the datasets}:

    Q: \{question\}
    
    A: Let's think step by step.
\end{tcolorbox}
\caption{Prompt of Zero-shot-CoT (i.e., ZeroCoT) for all datasets.}
\label{prompt_zero_cot}
\end{figure}

% \subsubsection{Plan-and-solve (PS)}
\begin{figure}[h]
\begin{tcolorbox}
\small
    % \textbf{Plan-and-solve (PS): Prompt for all the arithmetic reasoning datasets}:
    
    Q: \{question\}
    
    A: Let's first understand the problem, extract relevant variables and their corresponding numerals, and make and devise a complete plan. Then, let's carry out the plan, calculate intermediate variables (pay attention to correct numerical calculation and commonsense), solve the problem step by step, and show the answer.
\end{tcolorbox} 
\caption{Prompt of Plan-and-solve (i.e., PS) for all the arithmetic reasoning datasets (including MATH dataset).}
\label{prompt_PS_arithmetic}
\end{figure}

\begin{figure} 
\begin{tcolorbox}
\small
    % \textbf{Plan-and-solve (PS): Prompt for the commonsense reasoning dataset CSQA}:
    
    Q: \{question\}
    
    A: Let’s first prepare relevant information and make a plan. Then, let’s answer the question step by step (pay attention to commonsense and logical coherence).
\end{tcolorbox} 
\caption{Prompt of Plan-and-solve (i.e., PS) for the commonsense reasoning dataset CSQA.}
\label{prompt_PS_csqa}
\end{figure}

\begin{figure} \begin{tcolorbox}
\small
    % \textbf{Plan-and-solve (PS): Prompt for the symbolic reasoning dataset LLC}:
    
    Q: \{question\}
    
    A: Let’s devise a plan and solve the problem step by step.
\end{tcolorbox} 
\caption{Prompt of Plan-and-solve (i.e., PS) for the symbolic reasoning dataset LLC.}
\label{prompt_PS_llc}
\end{figure}

% \subsubsection{Chain-of-thought (CoT) \& COT\_SC}
\begin{figure}[h] 
\begin{tcolorbox}
\small
% \textbf{Chain-of-thought (CoT) \& COT\_SC: Four-shot exemplars for all the mathematical reasoning datasets excluding the AQuA dataset}:

Q: Four years ago, Kody was only half as old as Mohamed. If Mohamed is currently twice 30 years old, how old is Kody?

A: We were told that Mohamed is currently twice 30 years old, so he is currently 30 * 2 = 60 years old. That means that four years ago he must have been 60 - 4 = 56 years old. Four years ago, Kody was half as old as Mohamed, so Kody must have been 56 / 2 = 28 years old then. Since Kody was 28 years old four years ago, she must now be 28 + 4 = 32 years old. The answer is 32.

\dots

Q: \{question\}

A:
\end{tcolorbox} 
\caption{Prompt of CoT \& COT\_SC for all the mathematical reasoning datasets excluding the AQuA dataset}
\label{prompt_cot_math}
\end{figure}

\begin{figure}[h] 
\begin{tcolorbox}
    \small
    % \textbf{Chain-of-thought (CoT) \& COT\_SC: Four-shot exemplars for the commonsense reasoning dataset CSQA}:
    
Q: What do people use to absorb extra ink from a fountain pen? Answer Choices: (a) shirt pocket (b) calligrapher’s hand (c) inkwell (d) desk drawer (e) blotter

A: The answer must be an item that can absorb ink. Of the above choices, only blotters are used to absorb ink. So the answer is (e).

\dots

Q: \{question\}

A:
\end{tcolorbox} 
\caption{Prompt of CoT \& COT\_SC for the commonsense reasoning dataset CSQA.}
\label{prompt_cot_csqa}
\end{figure}

\begin{figure}[h] 
\begin{tcolorbox}
    \small
    % \textbf{Chain-of-thought (CoT) \& COT\_SC: Four-shot exemplars for the symbolic reasoning dataset LLC}:

Q: Take the last letters of the words in "Elon Musk" and concatenate them.

A: The last letter of "Elon" is "n". The last letter of "Musk" is "k". Concatenating them is "nk". The answer is "nk".

\dots

Q: \{question\}

A:
\end{tcolorbox} 
\caption{Prompt of CoT \& COT\_SC for the symbolic reasoning dataset LLC.}
\label{prompt_cot_llc}
\end{figure}

% \subsubsection{Least-to-most (L2M)}
\begin{figure}[h]
\begin{tcolorbox}
\small
% \textbf{Least-to-most (L2M): Four-shot exemplars for all the mathematical reasoning datasets excluding the AQuA dataset}:

Q: Four years ago, Kody was only half as old as Mohamed. If Mohamed is currently twice 30 years old, how old is Kody?

A: Let’s break down this problem: 1. How old is Mohamed currently? 2. How old was Mohamed four years ago? 3. How old was Kody four years ago? 4. How old is Kody?

1. We were told that Mohamed is currently twice 30 years old, so he is currently 30 * 2 = 60 years old. The answer is 60.
2. That means that four years ago he must have been 60 - 4 = 56 years old. The answer is 56.
3. Four years ago, Kody was half as old as Mohamed, so Kody must have been 56 / 2 = 28 years old then. The answer is 28.
4. Since Kody was 28 years old four years ago, she must now be 28 + 4 = 32 years old. The answer is 32.

\dots

Q: \{question\}

A: Let’s break down this problem:

\end{tcolorbox} 
\caption{Prompt of Least-to-most (i.e., L2M) for all the mathematical reasoning datasets excluding the AQuA dataset.}
\label{prompt_l2m_math}
\end{figure}

\begin{figure}[h]
\begin{tcolorbox}
\small
% \textbf{Least-to-most (L2M): Four-shot exemplars for the LLC dataset}:

Q: Take the last letters of the words in "think machine" and concatenate them.
A: Create sequential sublists of the list "think machine":
1. "think"
2. "think machine"
Concatenate the last letters of the words within each sublist sequentially:
1. "think": The last letter of "think" is "k".
2. "think machine": "think" outputs "k". The last letter of "machine" is "e". Concatenating "k", "e" leads to "ke".
The answer is "ke".

\dots

Q: \{question\}
A: Create sequential sublists of the list

\end{tcolorbox} 
\caption{Prompt of Least-to-most (i.e., L2M) for the LLC dataset.}
\label{prompt_l2m_llc}
\end{figure}

\subsubsection{L2M's variants for decomposition granularity adaptation}
\label{Appendix_prompt_l2m_variants}
The following three prompts mainly differ from the decomposition granularity.
For example, facing the same problem, the prompt (L2M, $d_1$) may break it down into 2-3 sub-questions, the prompt (L2M, $d_2$) may decompose it into 4-5 sub-questions, and the prompt (L2M, $d_3$) may decompose it into 6-8 sub-questions. In addition, the difference between them and L2M lies in: L2M lacks precise control over decomposition granularity in its demonstrations, leading to a blend of various granularities. Conversely, in the demonstrations of these variants, the decomposition granularity is either coarse, medium, or fine, depending on the specific variant.

% \textbf{(1) The prompts of (L2M, \textit{coarse})}

\begin{figure}[h]
\begin{tcolorbox}
\small
% \textbf{(L2M, \textit{coarse}): Four-shot exemplars for all the mathematical reasoning datasets excluding the AQuA dataset}:

Q: Cappuccinos cost \$2, iced teas cost \$3, cafe lattes cost \$1.5 and espressos cost \$1 each. Sandy orders some drinks for herself and some friends. She orders three cappuccinos, two iced teas, two cafe lattes, and two espressos. How much change does she receive back for a twenty-dollar bill?

A: Let’s break down this problem: 1. How much did Sandy spend on drinks? 2. How much change does she receive back for a twenty-dollar bill?

1. Sandy ordered three cappuccinos, which cost \$2 each, so she spent \$2 * 3 = \$6 on cappuccinos. She ordered two iced teas, which cost \$3 each, so she spent \$3 * 2 = \$6 dollars on ice teas. She ordered two cafe lattes, which cost \$1.5 each, so she spent \$1.5 * 2 = \$3 on cafe lattes. She ordered two espressos, which cost \$1 each, so she spent \$1 * 2 = \$2 on espressos. So altogether, Sandy spent \$6 + \$6 + \$3 + \$2 = \$17 on drinks. The answer is 17.
2. Sandy will get \$20 - \$17 = \$3 as change. The answer is 3.

\dots

Q: \{question\}

A: Let’s break down this problem:

\end{tcolorbox} 
\caption{Prompt of (L2M, \textit{coarse}) for all the mathematical reasoning datasets excluding the AQuA dataset.}
\label{prompt_l2m_coarse_math}
\end{figure}

% \textbf{(2) The prompts of (L2M, \textit{medium})}

\begin{figure} [h]
\begin{tcolorbox}
\small
% \textbf{(L2M, \textit{medium}): Four-shot exemplars for all the mathematical reasoning datasets excluding the AQuA dataset}:

Q: Cappuccinos cost \$2, iced teas cost \$3, cafe lattes cost \$1.5 and espressos cost \$1 each. Sandy orders some drinks for herself and some friends. She orders three cappuccinos, two iced teas, two cafe lattes, and two espressos. How much change does she receive back for a twenty-dollar bill?

A: Let’s break down this problem: 1. How much did the cappuccinos cost in total? 2. How much did the iced teas cost in total? 3. How much did the cafe lattes cost in total? 4. How much did the espressos cost in total? 5. How much did Sandy spend on drinks? 6. How much change does she receive back for a twenty-dollar bill?

1. Sandy ordered three cappuccinos, which cost \$2 each, so she spent \$2 * 3 = \$6 on cappuccinos. The answer is 6.
2. She ordered two iced teas, which cost \$3 each, so she spent \$3 * 2 = \$6 dollars on ice teas. The answer is 6.
3. She ordered two cafe lattes, which cost \$1.5 each, so she spent \$1.5 * 2 = \$3 on cafe lattes. The answer is 3.
4. She ordered two espressos, which cost \$1 each, so she spent \$1 * 2 = \$2 on espressos. The answer is 2.
5. So altogether, Sandy spent \$6 + \$6 + \$3 + \$2 = \$17 on drinks. The answer is 17.
6. Sandy will get \$20 - \$17 = \$3 as change. The answer is 3.

\dots

Q: \{question\}

A: Let’s break down this problem:
\end{tcolorbox} 
\caption{Prompt of (L2M, \textit{medium}) for all the mathematical reasoning datasets excluding the AQuA dataset.}
\label{prompt_l2m_medium_math}
\end{figure}

% \textbf{(3) The prompts of (L2M, \textit{fine})}

\begin{figure}[h]
\begin{tcolorbox}
\small
\textbf{(L2M, \textit{fine}): Four-shot exemplars for all the mathematical reasoning datasets excluding the AQuA dataset}:

Q: Cappuccinos cost \$2, iced teas cost \$3, cafe lattes cost \$1.5 and espressos cost \$1 each. Sandy orders some drinks for herself and some friends. She orders three cappuccinos, two iced teas, two cafe lattes, and two espressos. How much change does she receive back for a twenty-dollar bill?

A: Let’s break down this problem: 1. How many cappuccinos did Sandy order? 2. How much did the cappuccinos cost in total? 3. How many iced teas did Sandy order? 4. How much did the iced teas cost in total? 5. How many cafe lattes did Sandy order? 6. How much did the cafe lattes cost in total? 7. How many espressos did Sandy order? 8. How much did the espressos cost in total? 9. How much did Sandy spend on all drinks in total? 10. How much change does she receive back for a twenty-dollar bill?

1. Sandy ordered three cappuccinos. The answer is 3.
2. Each cappuccino cost \$2 each, so she spent \$2 * 3 = \$6 on cappuccinos. The answer is 6.
3. She ordered two iced teas. The answer is 2.
4. Each iced tea cost \$3 each, so she spent \$3 * 2 = \$6 dollars on ice teas. The answer is 6.
5. She ordered two cafe lattes. The answer is 2.
6. Each cafe latte cost \$1.5 each, so she spent \$1.5 * 2 = \$3 on cafe lattes. The answer is 3.
7. She ordered two espressos. The answer is 2.
8. Each espressos cost \$1 each, so she spent \$1 * 2 = \$2 on espressos. The answer is 2.
9. So altogether, Sandy spent \$6 + \$6 + \$3 + \$2 = \$17 on drinks. The answer is 17.
10. Sandy will get \$20 - \$17 = \$3 as change. The answer is 3.

Q: \{question\}

A: Let’s break down this problem:

\end{tcolorbox} 
\caption{Prompt of (L2M, \textit{fine}) for all the mathematical reasoning datasets excluding the AQuA dataset.}
\label{prompt_l2m_fine_math}
\end{figure}

% \verb+\printcredits+ command is used after appendix sections to list 
% author credit taxonomy contribution roles tagged using \verb+\credit+ 
% in frontmatter.

\printcredits

% \hypersetup{citecolor=blue}

% \bibliographystyle{cas-model2-names}
% \bibliographystyle{apacite}
\bibliographystyle{model5-names}


\bibliography{references}

\begin{thebibliography}{53}
\expandafter\ifx\csname natexlab\endcsname\relax\def\natexlab#1{#1}\fi
\providecommand{\url}[1]{\texttt{#1}}
\providecommand{\href}[2]{#2}
\providecommand{\path}[1]{#1}
\providecommand{\DOIprefix}{doi:}
\providecommand{\ArXivprefix}{arXiv:}
\providecommand{\URLprefix}{URL: }
\providecommand{\Pubmedprefix}{pmid:}
\providecommand{\doi}[1]{\href{http://dx.doi.org/#1}{\path{#1}}}
\providecommand{\Pubmed}[1]{\href{pmid:#1}{\path{#1}}}
\providecommand{\bibinfo}[2]{#2}
\ifx\xfnm\relax \def\xfnm[#1]{\unskip,\space#1}\fi
%Type = Inproceedings
\bibitem[{Aggarwal et~al.(2023)Aggarwal, Yang \& Mausam}]{adaptive_consistency}
\bibinfo{author}{Aggarwal, A. M.~P.}, \bibinfo{author}{Yang, Y.}, \& \bibinfo{author}{Mausam} (\bibinfo{year}{2023}).
\newblock \bibinfo{title}{Let's sample step by step: Adaptive-consistency for efficient reasoning and coding with llms}.
\newblock In {\it \bibinfo{booktitle}{Proceedings of the 2023 Conference on Empirical Methods in Natural Language Processing, {EMNLP} 2023}\/} (pp. \bibinfo{pages}{12375--12396}).
%Type = Article
\bibitem[{Chen et~al.(2023{\natexlab{a}})Chen, Zaharia \& Zou}]{frugalgpt}
\bibinfo{author}{Chen, L.}, \bibinfo{author}{Zaharia, M.}, \& \bibinfo{author}{Zou, J.} (\bibinfo{year}{2023}{\natexlab{a}}).
\newblock \bibinfo{title}{Frugalgpt: How to use large language models while reducing cost and improving performance}.
\newblock {\it \bibinfo{journal}{CoRR}\/},  {\it \bibinfo{volume}{abs/2305.05176}\/}. \href{http://arxiv.org/abs/2305.05176}{\tt arXiv:2305.05176}.
%Type = Article
\bibitem[{Chen et~al.(2023{\natexlab{b}})Chen, Ma, Wang \& Cohen}]{PoT}
\bibinfo{author}{Chen, W.}, \bibinfo{author}{Ma, X.}, \bibinfo{author}{Wang, X.}, \& \bibinfo{author}{Cohen, W.~W.} (\bibinfo{year}{2023}{\natexlab{b}}).
\newblock \bibinfo{title}{Program of thoughts prompting: Disentangling computation from reasoning for numerical reasoning tasks}.
\newblock {\it \bibinfo{journal}{Trans. Mach. Learn. Res.}\/},  {\it \bibinfo{volume}{2023}\/}.
%Type = Misc
\bibitem[{Cobbe et~al.(2021)Cobbe, Kosaraju, Bavarian, Chen, Jun, Kaiser, Plappert, Tworek, Hilton, Nakano, Hesse \& Schulman}]{gsm8k}
\bibinfo{author}{Cobbe, K.}, \bibinfo{author}{Kosaraju, V.}, \bibinfo{author}{Bavarian, M.}, \bibinfo{author}{Chen, M.}, \bibinfo{author}{Jun, H.}, \bibinfo{author}{Kaiser, L.}, \bibinfo{author}{Plappert, M.}, \bibinfo{author}{Tworek, J.}, \bibinfo{author}{Hilton, J.}, \bibinfo{author}{Nakano, R.}, \bibinfo{author}{Hesse, C.}, \& \bibinfo{author}{Schulman, J.} (\bibinfo{year}{2021}).
\newblock \bibinfo{title}{Training verifiers to solve math word problems}.
\newblock \href{http://arxiv.org/abs/2110.14168}{\tt arXiv:2110.14168}.
%Type = Inproceedings
\bibitem[{Creswell et~al.(2023)Creswell, Shanahan \& Higgins}]{selection-inference}
\bibinfo{author}{Creswell, A.}, \bibinfo{author}{Shanahan, M.}, \& \bibinfo{author}{Higgins, I.} (\bibinfo{year}{2023}).
\newblock \bibinfo{title}{Selection-inference: Exploiting large language models for interpretable logical reasoning}.
\newblock In {\it \bibinfo{booktitle}{The Eleventh International Conference on Learning Representations, {ICLR} 2023}\/}.
%Type = Inproceedings
\bibitem[{Gao et~al.(2023)Gao, Madaan, Zhou, Alon, Liu, Yang, Callan \& Neubig}]{PAL}
\bibinfo{author}{Gao, L.}, \bibinfo{author}{Madaan, A.}, \bibinfo{author}{Zhou, S.}, \bibinfo{author}{Alon, U.}, \bibinfo{author}{Liu, P.}, \bibinfo{author}{Yang, Y.}, \bibinfo{author}{Callan, J.}, \& \bibinfo{author}{Neubig, G.} (\bibinfo{year}{2023}).
\newblock \bibinfo{title}{{PAL:} program-aided language models}.
\newblock In \bibinfo{editor}{A.~Krause}, \bibinfo{editor}{E.~Brunskill}, \bibinfo{editor}{K.~Cho}, \bibinfo{editor}{B.~Engelhardt}, \bibinfo{editor}{S.~Sabato}, \& \bibinfo{editor}{J.~Scarlett} (Eds.), {\it \bibinfo{booktitle}{International Conference on Machine Learning, {ICML} 2023}\/} (pp. \bibinfo{pages}{10764--10799}).
\newblock \bibinfo{publisher}{{PMLR}} volume \bibinfo{volume}{202} of {\it \bibinfo{series}{Proceedings of Machine Learning Research}\/}.
%Type = Inproceedings
\bibitem[{Gou et~al.(2024{\natexlab{a}})Gou, Shao, Gong, Shen, Yang, Duan \& Chen}]{critic}
\bibinfo{author}{Gou, Z.}, \bibinfo{author}{Shao, Z.}, \bibinfo{author}{Gong, Y.}, \bibinfo{author}{Shen, Y.}, \bibinfo{author}{Yang, Y.}, \bibinfo{author}{Duan, N.}, \& \bibinfo{author}{Chen, W.} (\bibinfo{year}{2024}{\natexlab{a}}).
\newblock \bibinfo{title}{{CRITIC:} large language models can self-correct with tool-interactive critiquing}.
\newblock In {\it \bibinfo{booktitle}{The Twelfth International Conference on Learning Representations, {ICLR} 2024}\/}.
%Type = Inproceedings
\bibitem[{Gou et~al.(2024{\natexlab{b}})Gou, Shao, Gong, Shen, Yang, Huang, Duan \& Chen}]{tora}
\bibinfo{author}{Gou, Z.}, \bibinfo{author}{Shao, Z.}, \bibinfo{author}{Gong, Y.}, \bibinfo{author}{Shen, Y.}, \bibinfo{author}{Yang, Y.}, \bibinfo{author}{Huang, M.}, \bibinfo{author}{Duan, N.}, \& \bibinfo{author}{Chen, W.} (\bibinfo{year}{2024}{\natexlab{b}}).
\newblock \bibinfo{title}{Tora: {A} tool-integrated reasoning agent for mathematical problem solving}.
\newblock In {\it \bibinfo{booktitle}{The Twelfth International Conference on Learning Representations, {ICLR} 2024}\/}.
%Type = Article
\bibitem[{He et~al.(2023)He, Zhang \& Roth}]{RR}
\bibinfo{author}{He, H.}, \bibinfo{author}{Zhang, H.}, \& \bibinfo{author}{Roth, D.} (\bibinfo{year}{2023}).
\newblock \bibinfo{title}{Rethinking with retrieval: Faithful large language model inference}.
\newblock {\it \bibinfo{journal}{CoRR}\/},  {\it \bibinfo{volume}{abs/2301.00303}\/}.
%Type = Inproceedings
\bibitem[{Hosseini et~al.(2014)Hosseini, Hajishirzi, Etzioni \& Kushman}]{addsub}
\bibinfo{author}{Hosseini, M.~J.}, \bibinfo{author}{Hajishirzi, H.}, \bibinfo{author}{Etzioni, O.}, \& \bibinfo{author}{Kushman, N.} (\bibinfo{year}{2014}).
\newblock \bibinfo{title}{Learning to solve arithmetic word problems with verb categorization}.
\newblock In {\it \bibinfo{booktitle}{Proceedings of the 2014 Conference on Empirical Methods in Natural Language Processing, {EMNLP} 2014}\/} (pp. \bibinfo{pages}{523--533}).
%Type = Inproceedings
\bibitem[{Hsieh et~al.(2023)Hsieh, Li, Yeh, Nakhost, Fujii, Ratner, Krishna, Lee \& Pfister}]{distill_step_by_step}
\bibinfo{author}{Hsieh, C.-Y.}, \bibinfo{author}{Li, C.-L.}, \bibinfo{author}{Yeh, C.-k.}, \bibinfo{author}{Nakhost, H.}, \bibinfo{author}{Fujii, Y.}, \bibinfo{author}{Ratner, A.}, \bibinfo{author}{Krishna, R.}, \bibinfo{author}{Lee, C.-Y.}, \& \bibinfo{author}{Pfister, T.} (\bibinfo{year}{2023}).
\newblock \bibinfo{title}{Distilling step-by-step! outperforming larger language models with less training data and smaller model sizes}.
\newblock In {\it \bibinfo{booktitle}{Findings of the Association for Computational Linguistics: ACL 2023}\/} (pp. \bibinfo{pages}{8003--8017}).
%Type = Inproceedings
\bibitem[{Huang \& Chang(2023)}]{survey-llm-reason}
\bibinfo{author}{Huang, J.}, \& \bibinfo{author}{Chang, K. C.-C.} (\bibinfo{year}{2023}).
\newblock \bibinfo{title}{Towards reasoning in large language models: A survey}.
\newblock In {\it \bibinfo{booktitle}{Findings of the Association for Computational Linguistics: ACL 2023}\/} (pp. \bibinfo{pages}{1049--1065}).
%Type = Inproceedings
\bibitem[{Jung et~al.(2022)Jung, Qin, Welleck, Brahman, Bhagavatula, Le~Bras \& Choi}]{maieutic}
\bibinfo{author}{Jung, J.}, \bibinfo{author}{Qin, L.}, \bibinfo{author}{Welleck, S.}, \bibinfo{author}{Brahman, F.}, \bibinfo{author}{Bhagavatula, C.}, \bibinfo{author}{Le~Bras, R.}, \& \bibinfo{author}{Choi, Y.} (\bibinfo{year}{2022}).
\newblock \bibinfo{title}{Maieutic prompting: Logically consistent reasoning with recursive explanations}.
\newblock In {\it \bibinfo{booktitle}{Proceedings of the 2022 Conference on Empirical Methods in Natural Language Processing ({EMNLP})}\/} (pp. \bibinfo{pages}{1266--1279}).
%Type = Book
\bibitem[{Kahneman(2011)}]{daniel2017thinking}
\bibinfo{author}{Kahneman, D.} (\bibinfo{year}{2011}).
\newblock {\it \bibinfo{title}{Thinking, fast and slow}\/}.
\newblock \bibinfo{publisher}{Farrar, Straus and Giroux}.
%Type = Inproceedings
\bibitem[{Khot et~al.(2023)Khot, Trivedi, Finlayson, Fu, Richardson, Clark \& Sabharwal}]{Decomposed-Prompting}
\bibinfo{author}{Khot, T.}, \bibinfo{author}{Trivedi, H.}, \bibinfo{author}{Finlayson, M.}, \bibinfo{author}{Fu, Y.}, \bibinfo{author}{Richardson, K.}, \bibinfo{author}{Clark, P.}, \& \bibinfo{author}{Sabharwal, A.} (\bibinfo{year}{2023}).
\newblock \bibinfo{title}{Decomposed prompting: A modular approach for solving complex tasks}.
\newblock In {\it \bibinfo{booktitle}{The Eleventh International Conference on Learning Representations}\/}.
%Type = Inproceedings
\bibitem[{Kojima et~al.(2022)Kojima, Gu, Reid, Matsuo \& Iwasawa}]{ZeroCoT}
\bibinfo{author}{Kojima, T.}, \bibinfo{author}{Gu, S.~S.}, \bibinfo{author}{Reid, M.}, \bibinfo{author}{Matsuo, Y.}, \& \bibinfo{author}{Iwasawa, Y.} (\bibinfo{year}{2022}).
\newblock \bibinfo{title}{Large language models are zero-shot reasoners}.
\newblock In {\it \bibinfo{booktitle}{Advances in Neural Information Processing Systems}\/}.
%Type = Article
\bibitem[{Koncel-Kedziorski et~al.(2015)Koncel-Kedziorski, Hajishirzi, Sabharwal, Etzioni \& Ang}]{singleeq}
\bibinfo{author}{Koncel-Kedziorski, R.}, \bibinfo{author}{Hajishirzi, H.}, \bibinfo{author}{Sabharwal, A.}, \bibinfo{author}{Etzioni, O.}, \& \bibinfo{author}{Ang, S.~D.} (\bibinfo{year}{2015}).
\newblock \bibinfo{title}{Parsing algebraic word problems into equations}.
\newblock {\it \bibinfo{journal}{Transactions of the Association for Computational Linguistics}\/},  {\it \bibinfo{volume}{3}\/}, \bibinfo{pages}{585--597}.
%Type = Inproceedings
\bibitem[{Li et~al.(2019)Li, Wang, Zhang, Wang, Dai \& Zhang}]{group-attn}
\bibinfo{author}{Li, J.}, \bibinfo{author}{Wang, L.}, \bibinfo{author}{Zhang, J.}, \bibinfo{author}{Wang, Y.}, \bibinfo{author}{Dai, B.~T.}, \& \bibinfo{author}{Zhang, D.} (\bibinfo{year}{2019}).
\newblock \bibinfo{title}{Modeling intra-relation in math word problems with different functional multi-head attentions}.
\newblock In {\it \bibinfo{booktitle}{Proceedings of the 57th Annual Meeting of the Association for Computational Linguistics}\/} (pp. \bibinfo{pages}{6162--6167}).
%Type = Article
\bibitem[{Liang et~al.(2024)Liang, Wang, Zhong, Wang, Li, Jia \& Wan}]{medical_reasoning}
\bibinfo{author}{Liang, X.}, \bibinfo{author}{Wang, D.}, \bibinfo{author}{Zhong, H.}, \bibinfo{author}{Wang, Q.}, \bibinfo{author}{Li, R.}, \bibinfo{author}{Jia, R.}, \& \bibinfo{author}{Wan, B.} (\bibinfo{year}{2024}).
\newblock \bibinfo{title}{Candidate-heuristic in-context learning: A new framework for enhancing medical visual question answering with llms}.
\newblock {\it \bibinfo{journal}{Information Processing and Management}\/},  {\it \bibinfo{volume}{61}\/}, \bibinfo{pages}{103805}.
%Type = Inproceedings
\bibitem[{Ling et~al.(2017{\natexlab{a}})Ling, Yogatama, Dyer \& Blunsom}]{program-induction}
\bibinfo{author}{Ling, W.}, \bibinfo{author}{Yogatama, D.}, \bibinfo{author}{Dyer, C.}, \& \bibinfo{author}{Blunsom, P.} (\bibinfo{year}{2017}{\natexlab{a}}).
\newblock \bibinfo{title}{Program induction by rationale generation: Learning to solve and explain algebraic word problems}.
\newblock In {\it \bibinfo{booktitle}{Proceedings of the 55th Annual Meeting of the Association for Computational Linguistics (Volume 1: Long Papers)}\/} (pp. \bibinfo{pages}{158--167}).
%Type = Inproceedings
\bibitem[{Ling et~al.(2017{\natexlab{b}})Ling, Yogatama, Dyer \& Blunsom}]{aqua}
\bibinfo{author}{Ling, W.}, \bibinfo{author}{Yogatama, D.}, \bibinfo{author}{Dyer, C.}, \& \bibinfo{author}{Blunsom, P.} (\bibinfo{year}{2017}{\natexlab{b}}).
\newblock \bibinfo{title}{Program induction by rationale generation: Learning to solve and explain algebraic word problems}.
\newblock In {\it \bibinfo{booktitle}{Proceedings of the 55th Annual Meeting of the Association for Computational Linguistics (Volume 1: Long Papers)}\/} (pp. \bibinfo{pages}{158--167}).
%Type = Inproceedings
\bibitem[{Lu et~al.(2023)Lu, Qiu, Yu, Welleck \& Chang}]{survey-dl-math}
\bibinfo{author}{Lu, P.}, \bibinfo{author}{Qiu, L.}, \bibinfo{author}{Yu, W.}, \bibinfo{author}{Welleck, S.}, \& \bibinfo{author}{Chang, K.-W.} (\bibinfo{year}{2023}).
\newblock \bibinfo{title}{A survey of deep learning for mathematical reasoning}.
\newblock In {\it \bibinfo{booktitle}{Proceedings of the 61st Annual Meeting of the Association for Computational Linguistics (Volume 1: Long Papers)}\/} (pp. \bibinfo{pages}{14605--14631}).
%Type = Inproceedings
\bibitem[{Madaan et~al.(2023)Madaan, Tandon, Gupta, Hallinan, Gao, Wiegreffe, Alon, Dziri, Prabhumoye, Yang, Gupta, Majumder, Hermann, Welleck, Yazdanbakhsh \& Clark}]{self-refine}
\bibinfo{author}{Madaan, A.}, \bibinfo{author}{Tandon, N.}, \bibinfo{author}{Gupta, P.}, \bibinfo{author}{Hallinan, S.}, \bibinfo{author}{Gao, L.}, \bibinfo{author}{Wiegreffe, S.}, \bibinfo{author}{Alon, U.}, \bibinfo{author}{Dziri, N.}, \bibinfo{author}{Prabhumoye, S.}, \bibinfo{author}{Yang, Y.}, \bibinfo{author}{Gupta, S.}, \bibinfo{author}{Majumder, B.~P.}, \bibinfo{author}{Hermann, K.}, \bibinfo{author}{Welleck, S.}, \bibinfo{author}{Yazdanbakhsh, A.}, \& \bibinfo{author}{Clark, P.} (\bibinfo{year}{2023}).
\newblock \bibinfo{title}{Self-refine: Iterative refinement with self-feedback}.
\newblock In {\it \bibinfo{booktitle}{Thirty-seventh Conference on Neural Information Processing Systems}\/}.
%Type = Article
\bibitem[{Pan et~al.(2023)Pan, Saxon, Xu, Nathani, Wang \& Wang}]{survey-auto-correct}
\bibinfo{author}{Pan, L.}, \bibinfo{author}{Saxon, M.}, \bibinfo{author}{Xu, W.}, \bibinfo{author}{Nathani, D.}, \bibinfo{author}{Wang, X.}, \& \bibinfo{author}{Wang, W.~Y.} (\bibinfo{year}{2023}).
\newblock \bibinfo{title}{Automatically correcting large language models: Surveying the landscape of diverse self-correction strategies}.
\newblock {\it \bibinfo{journal}{arXiv preprint arXiv:2308.03188}\/}, .
%Type = Inproceedings
\bibitem[{Patel et~al.(2021)Patel, Bhattamishra \& Goyal}]{svamp}
\bibinfo{author}{Patel, A.}, \bibinfo{author}{Bhattamishra, S.}, \& \bibinfo{author}{Goyal, N.} (\bibinfo{year}{2021}).
\newblock \bibinfo{title}{Are {NLP} models really able to solve simple math word problems?}
\newblock In {\it \bibinfo{booktitle}{Proceedings of the 2021 Conference of the North American Chapter of the Association for Computational Linguistics: Human Language Technologies}\/} (pp. \bibinfo{pages}{2080--2094}).
%Type = Inproceedings
\bibitem[{Qiao et~al.(2023)Qiao, Ou, Zhang, Chen, Yao, Deng, Tan, Huang \& Chen}]{survey-llm-prompt-reason}
\bibinfo{author}{Qiao, S.}, \bibinfo{author}{Ou, Y.}, \bibinfo{author}{Zhang, N.}, \bibinfo{author}{Chen, X.}, \bibinfo{author}{Yao, Y.}, \bibinfo{author}{Deng, S.}, \bibinfo{author}{Tan, C.}, \bibinfo{author}{Huang, F.}, \& \bibinfo{author}{Chen, H.} (\bibinfo{year}{2023}).
\newblock \bibinfo{title}{Reasoning with language model prompting: A survey}.
\newblock In {\it \bibinfo{booktitle}{Proceedings of the 61st Annual Meeting of the Association for Computational Linguistics (Volume 1: Long Papers)}\/} (pp. \bibinfo{pages}{5368--5393}).
\newblock \bibinfo{address}{Toronto, Canada}: \bibinfo{publisher}{Association for Computational Linguistics}.
%Type = Article
\bibitem[{Qiu et~al.(2024)Qiu, Xie, Liu \& Hu}]{multimodal_reasoning}
\bibinfo{author}{Qiu, C.}, \bibinfo{author}{Xie, Z.}, \bibinfo{author}{Liu, M.}, \& \bibinfo{author}{Hu, H.} (\bibinfo{year}{2024}).
\newblock \bibinfo{title}{Explainable knowledge reasoning via thought chains for knowledge-based visual question answering}.
\newblock {\it \bibinfo{journal}{Information Processing and Management}\/},  {\it \bibinfo{volume}{61}\/}, \bibinfo{pages}{103726}.
%Type = Inproceedings
\bibitem[{Roy \& Roth(2015)}]{multiarith}
\bibinfo{author}{Roy, S.}, \& \bibinfo{author}{Roth, D.} (\bibinfo{year}{2015}).
\newblock \bibinfo{title}{Solving general arithmetic word problems}.
\newblock In {\it \bibinfo{booktitle}{Proceedings of the 2015 Conference on Empirical Methods in Natural Language Processing ({EMNLP})}\/} (pp. \bibinfo{pages}{1743--1752}).
%Type = Inproceedings
\bibitem[{Shen et~al.(2021)Shen, Yin, Li, Shang, Jiang, Zhang \& Liu}]{generate-rank}
\bibinfo{author}{Shen, J.}, \bibinfo{author}{Yin, Y.}, \bibinfo{author}{Li, L.}, \bibinfo{author}{Shang, L.}, \bibinfo{author}{Jiang, X.}, \bibinfo{author}{Zhang, M.}, \& \bibinfo{author}{Liu, Q.} (\bibinfo{year}{2021}).
\newblock \bibinfo{title}{Generate {\&} rank: A multi-task framework for math word problems}.
\newblock In {\it \bibinfo{booktitle}{Findings of the Association for Computational Linguistics: EMNLP 2021}\/} (pp. \bibinfo{pages}{2269--2279}).
%Type = Article
\bibitem[{Sloman(1996)}]{sloman1996empirical}
\bibinfo{author}{Sloman, S.~A.} (\bibinfo{year}{1996}).
\newblock \bibinfo{title}{The empirical case for two systems of reasoning.}
\newblock {\it \bibinfo{journal}{Psychological bulletin}\/},  {\it \bibinfo{volume}{119}\/}, \bibinfo{pages}{3}.
%Type = Article
\bibitem[{Snell et~al.(2024)Snell, Lee, Xu \& Kumar}]{scaling_test_time}
\bibinfo{author}{Snell, C.}, \bibinfo{author}{Lee, J.}, \bibinfo{author}{Xu, K.}, \& \bibinfo{author}{Kumar, A.} (\bibinfo{year}{2024}).
\newblock \bibinfo{title}{Scaling llm test-time compute optimally can be more effective than scaling model parameters}.
\newblock {\it \bibinfo{journal}{arXiv preprint arXiv:2408.03314}\/}, .
%Type = Inproceedings
\bibitem[{Talmor et~al.(2019)Talmor, Herzig, Lourie \& Berant}]{CSQA}
\bibinfo{author}{Talmor, A.}, \bibinfo{author}{Herzig, J.}, \bibinfo{author}{Lourie, N.}, \& \bibinfo{author}{Berant, J.} (\bibinfo{year}{2019}).
\newblock \bibinfo{title}{{C}ommonsense{QA}: A question answering challenge targeting commonsense knowledge}.
\newblock In {\it \bibinfo{booktitle}{Proceedings of the 2019 Conference of the North {A}merican Chapter of the Association for Computational Linguistics: Human Language Technologies, Volume 1 (Long and Short Papers)}\/} (pp. \bibinfo{pages}{4149--4158}).
%Type = Inproceedings
\bibitem[{Wang et~al.(2018)Wang, Wang, Cai, Zhang \& Liu}]{MATH-EN}
\bibinfo{author}{Wang, L.}, \bibinfo{author}{Wang, Y.}, \bibinfo{author}{Cai, D.}, \bibinfo{author}{Zhang, D.}, \& \bibinfo{author}{Liu, X.} (\bibinfo{year}{2018}).
\newblock \bibinfo{title}{Translating a math word problem to a expression tree}.
\newblock In {\it \bibinfo{booktitle}{Proceedings of the 2018 Conference on Empirical Methods in Natural Language Processing}\/} (pp. \bibinfo{pages}{1064--1069}).
%Type = Inproceedings
\bibitem[{Wang et~al.(2023{\natexlab{a}})Wang, Xu, Lan, Hu, Lan, Lee \& Lim}]{Plan-and-solve}
\bibinfo{author}{Wang, L.}, \bibinfo{author}{Xu, W.}, \bibinfo{author}{Lan, Y.}, \bibinfo{author}{Hu, Z.}, \bibinfo{author}{Lan, Y.}, \bibinfo{author}{Lee, R. K.-W.}, \& \bibinfo{author}{Lim, E.-P.} (\bibinfo{year}{2023}{\natexlab{a}}).
\newblock \bibinfo{title}{Plan-and-solve prompting: Improving zero-shot chain-of-thought reasoning by large language models}.
\newblock In {\it \bibinfo{booktitle}{Proceedings of the 61st Annual Meeting of the Association for Computational Linguistics (Volume 1: Long Papers)}\/} (pp. \bibinfo{pages}{2609--2634}).
%Type = Article
\bibitem[{Wang et~al.(2023{\natexlab{b}})Wang, Yu, Tan, O'Brien, Pasunuru, Dwivedi-Yu, Golovneva, Zettlemoyer, Fazel-Zarandi \& Celikyilmaz}]{shepherd}
\bibinfo{author}{Wang, T.}, \bibinfo{author}{Yu, P.}, \bibinfo{author}{Tan, X.~E.}, \bibinfo{author}{O'Brien, S.}, \bibinfo{author}{Pasunuru, R.}, \bibinfo{author}{Dwivedi-Yu, J.}, \bibinfo{author}{Golovneva, O.}, \bibinfo{author}{Zettlemoyer, L.}, \bibinfo{author}{Fazel-Zarandi, M.}, \& \bibinfo{author}{Celikyilmaz, A.} (\bibinfo{year}{2023}{\natexlab{b}}).
\newblock \bibinfo{title}{Shepherd: A critic for language model generation}.
\newblock {\it \bibinfo{journal}{arXiv preprint arXiv:2308.04592}\/}, .
%Type = Inproceedings
\bibitem[{Wang et~al.(2023{\natexlab{c}})Wang, Wei, Schuurmans, Le, Chi, Narang, Chowdhery \& Zhou}]{SC}
\bibinfo{author}{Wang, X.}, \bibinfo{author}{Wei, J.}, \bibinfo{author}{Schuurmans, D.}, \bibinfo{author}{Le, Q.~V.}, \bibinfo{author}{Chi, E.~H.}, \bibinfo{author}{Narang, S.}, \bibinfo{author}{Chowdhery, A.}, \& \bibinfo{author}{Zhou, D.} (\bibinfo{year}{2023}{\natexlab{c}}).
\newblock \bibinfo{title}{Self-consistency improves chain of thought reasoning in language models}.
\newblock In {\it \bibinfo{booktitle}{The Eleventh International Conference on Learning Representations}\/}.
%Type = Inproceedings
\bibitem[{Wang et~al.(2017)Wang, Liu \& Shi}]{DNS}
\bibinfo{author}{Wang, Y.}, \bibinfo{author}{Liu, X.}, \& \bibinfo{author}{Shi, S.} (\bibinfo{year}{2017}).
\newblock \bibinfo{title}{Deep neural solver for math word problems}.
\newblock In {\it \bibinfo{booktitle}{Proceedings of the 2017 conference on empirical methods in natural language processing}\/} (pp. \bibinfo{pages}{845--854}).
%Type = Inproceedings
\bibitem[{Wang et~al.(2023{\natexlab{d}})Wang, Huang, Liu, Wang, Song, Zhang, Huang, Wei, Deng, Sun \& Zhang}]{tailored_learning}
\bibinfo{author}{Wang, Z.}, \bibinfo{author}{Huang, S.}, \bibinfo{author}{Liu, Y.}, \bibinfo{author}{Wang, J.}, \bibinfo{author}{Song, M.}, \bibinfo{author}{Zhang, Z.}, \bibinfo{author}{Huang, H.}, \bibinfo{author}{Wei, F.}, \bibinfo{author}{Deng, W.}, \bibinfo{author}{Sun, F.}, \& \bibinfo{author}{Zhang, Q.} (\bibinfo{year}{2023}{\natexlab{d}}).
\newblock \bibinfo{title}{Democratizing reasoning ability: Tailored learning from large language model}.
\newblock In {\it \bibinfo{booktitle}{The 2023 Conference on Empirical Methods in Natural Language Processing}\/}.
%Type = Inproceedings
\bibitem[{Wei et~al.(2022)Wei, Wang, Schuurmans, Bosma, brian ichter, Xia, Chi, Le \& Zhou}]{CoT}
\bibinfo{author}{Wei, J.}, \bibinfo{author}{Wang, X.}, \bibinfo{author}{Schuurmans, D.}, \bibinfo{author}{Bosma, M.}, \bibinfo{author}{brian ichter}, \bibinfo{author}{Xia, F.}, \bibinfo{author}{Chi, E.~H.}, \bibinfo{author}{Le, Q.~V.}, \& \bibinfo{author}{Zhou, D.} (\bibinfo{year}{2022}).
\newblock \bibinfo{title}{Chain of thought prompting elicits reasoning in large language models}.
\newblock In {\it \bibinfo{booktitle}{Advances in Neural Information Processing Systems}\/}.
%Type = Inproceedings
\bibitem[{Weng et~al.(2023)Weng, Zhu, Xia, Li, He, Liu \& Zhao}]{Self-Verification}
\bibinfo{author}{Weng, Y.}, \bibinfo{author}{Zhu, M.}, \bibinfo{author}{Xia, F.}, \bibinfo{author}{Li, B.}, \bibinfo{author}{He, S.}, \bibinfo{author}{Liu, K.}, \& \bibinfo{author}{Zhao, J.} (\bibinfo{year}{2023}).
\newblock \bibinfo{title}{Large language models are better reasoners with self-verification}.
\newblock In {\it \bibinfo{booktitle}{The 2023 Conference on Empirical Methods in Natural Language Processing ({EMNLP})}\/} (p. \bibinfo{pages}{2550–2575}).
%Type = Inproceedings
\bibitem[{Wu et~al.(2020)Wu, Zhang, Fu \& Huang}]{KA-S2T}
\bibinfo{author}{Wu, Q.}, \bibinfo{author}{Zhang, Q.}, \bibinfo{author}{Fu, J.}, \& \bibinfo{author}{Huang, X.} (\bibinfo{year}{2020}).
\newblock \bibinfo{title}{A knowledge-aware sequence-to-tree network for math word problem solving}.
\newblock In {\it \bibinfo{booktitle}{Proceedings of the 2020 Conference on Empirical Methods in Natural Language Processing (EMNLP)}\/} (pp. \bibinfo{pages}{7137--7146}).
%Type = Article
\bibitem[{Wu et~al.(2024)Wu, Sun, Li, Welleck \& Yang}]{empirical}
\bibinfo{author}{Wu, Y.}, \bibinfo{author}{Sun, Z.}, \bibinfo{author}{Li, S.}, \bibinfo{author}{Welleck, S.}, \& \bibinfo{author}{Yang, Y.} (\bibinfo{year}{2024}).
\newblock \bibinfo{title}{An empirical analysis of compute-optimal inference for problem-solving with language models}.
\newblock {\it \bibinfo{journal}{arXiv preprint arXiv:2408.00724}\/}, .
%Type = Article
\bibitem[{Xiao et~al.(2023)Xiao, Huang, Song \& Tang}]{RGFNet}
\bibinfo{author}{Xiao, J.}, \bibinfo{author}{Huang, L.}, \bibinfo{author}{Song, Y.}, \& \bibinfo{author}{Tang, N.} (\bibinfo{year}{2023}).
\newblock \bibinfo{title}{A recursive tree-structured neural network with goal forgetting and information aggregation for solving math word problems}.
\newblock {\it \bibinfo{journal}{Information Processing and Management}\/},  {\it \bibinfo{volume}{60}\/}, \bibinfo{pages}{103324}.
%Type = Inproceedings
\bibitem[{Xie et~al.(2023)Xie, Kawaguchi, Zhao, Zhao, Kan, He \& Xie}]{SelfEval-Guided-Decoding}
\bibinfo{author}{Xie, Y.}, \bibinfo{author}{Kawaguchi, K.}, \bibinfo{author}{Zhao, Y.}, \bibinfo{author}{Zhao, X.}, \bibinfo{author}{Kan, M.-Y.}, \bibinfo{author}{He, J.}, \& \bibinfo{author}{Xie, Q.} (\bibinfo{year}{2023}).
\newblock \bibinfo{title}{Self-evaluation guided beam search for reasoning}.
\newblock In {\it \bibinfo{booktitle}{Thirty-seventh Conference on Neural Information Processing Systems}\/}.
%Type = Inproceedings
\bibitem[{Xie \& Sun(2019)}]{GTS}
\bibinfo{author}{Xie, Z.}, \& \bibinfo{author}{Sun, S.} (\bibinfo{year}{2019}).
\newblock \bibinfo{title}{A goal-driven tree-structured neural model for math word problems}.
\newblock In {\it \bibinfo{booktitle}{Proceedings of the Twenty-Eighth International Joint Conference on Artificial Intelligence, {IJCAI-19}}\/} (pp. \bibinfo{pages}{5299--5305}).
%Type = Inproceedings
\bibitem[{Yao et~al.(2023)Yao, Yu, Zhao, Shafran, Griffiths, Cao \& Narasimhan}]{ToT}
\bibinfo{author}{Yao, S.}, \bibinfo{author}{Yu, D.}, \bibinfo{author}{Zhao, J.}, \bibinfo{author}{Shafran, I.}, \bibinfo{author}{Griffiths, T.~L.}, \bibinfo{author}{Cao, Y.}, \& \bibinfo{author}{Narasimhan, K.~R.} (\bibinfo{year}{2023}).
\newblock \bibinfo{title}{Tree of thoughts: Deliberate problem solving with large language models}.
\newblock In {\it \bibinfo{booktitle}{Thirty-seventh Conference on Neural Information Processing Systems}\/}.
%Type = Article
\bibitem[{Yu et~al.(2023)Yu, Zhang, Liang, Jiang \& Sabharwal}]{Retrieval-Feedback}
\bibinfo{author}{Yu, W.}, \bibinfo{author}{Zhang, Z.}, \bibinfo{author}{Liang, Z.}, \bibinfo{author}{Jiang, M.}, \& \bibinfo{author}{Sabharwal, A.} (\bibinfo{year}{2023}).
\newblock \bibinfo{title}{Improving language models via plug-and-play retrieval feedback}.
\newblock {\it \bibinfo{journal}{arXiv preprint arXiv:2305.14002}\/}, .
%Type = Inproceedings
\bibitem[{Yue et~al.(2024)Yue, Zhao, Zhang, Du \& Yao}]{model_cascade}
\bibinfo{author}{Yue, M.}, \bibinfo{author}{Zhao, J.}, \bibinfo{author}{Zhang, M.}, \bibinfo{author}{Du, L.}, \& \bibinfo{author}{Yao, Z.} (\bibinfo{year}{2024}).
\newblock \bibinfo{title}{Large language model cascades with mixture of thought representations for cost-efficient reasoning}.
\newblock In {\it \bibinfo{booktitle}{The Twelfth International Conference on Learning Representations}\/}.
%Type = Inproceedings
\bibitem[{Zhang et~al.(2020)Zhang, Wang, Lee, Bin, Wang, Shao \& Lim}]{g2t-z}
\bibinfo{author}{Zhang, J.}, \bibinfo{author}{Wang, L.}, \bibinfo{author}{Lee, R. K.-W.}, \bibinfo{author}{Bin, Y.}, \bibinfo{author}{Wang, Y.}, \bibinfo{author}{Shao, J.}, \& \bibinfo{author}{Lim, E.-P.} (\bibinfo{year}{2020}).
\newblock \bibinfo{title}{Graph-to-tree learning for solving math word problems}.
\newblock In {\it \bibinfo{booktitle}{Proceedings of the 58th Annual Meeting of the Association for Computational Linguistics}\/} (pp. \bibinfo{pages}{3928--3937}).
%Type = Article
\bibitem[{Zhang et~al.(2022)Zhang, Zhou, Xie \& Huang}]{Hgen}
\bibinfo{author}{Zhang, Y.}, \bibinfo{author}{Zhou, G.}, \bibinfo{author}{Xie, Z.}, \& \bibinfo{author}{Huang, J.~X.} (\bibinfo{year}{2022}).
\newblock \bibinfo{title}{Hgen: Learning hierarchical heterogeneous graph encoding for math word problem solving}.
\newblock {\it \bibinfo{journal}{IEEE/ACM Transactions on Audio, Speech, and Language Processing}\/},  {\it \bibinfo{volume}{30}\/}, \bibinfo{pages}{816--828}.
%Type = Article
\bibitem[{Zhang et~al.(2024)Zhang, Zhou, Xie \& Huang}]{NERHRT}
\bibinfo{author}{Zhang, Y.}, \bibinfo{author}{Zhou, G.}, \bibinfo{author}{Xie, Z.}, \& \bibinfo{author}{Huang, J.~X.} (\bibinfo{year}{2024}).
\newblock \bibinfo{title}{Number-enhanced representation with hierarchical recursive tree decoding for math word problem solving}.
\newblock {\it \bibinfo{journal}{Information Processing and Management}\/},  {\it \bibinfo{volume}{61}\/}, \bibinfo{pages}{103585}.
%Type = Article
\bibitem[{Zheng et~al.(2023)Zheng, Liu, Xie, Li \& Li}]{PHP}
\bibinfo{author}{Zheng, C.}, \bibinfo{author}{Liu, Z.}, \bibinfo{author}{Xie, E.}, \bibinfo{author}{Li, Z.}, \& \bibinfo{author}{Li, Y.} (\bibinfo{year}{2023}).
\newblock \bibinfo{title}{Progressive-hint prompting improves reasoning in large language models}.
\newblock {\it \bibinfo{journal}{arXiv preprint arXiv:2304.09797}\/}, .
%Type = Inproceedings
\bibitem[{Zhou et~al.(2023)Zhou, Sch{\"a}rli, Hou, Wei, Scales, Wang, Schuurmans, Cui, Bousquet, Le \& Chi}]{L2M}
\bibinfo{author}{Zhou, D.}, \bibinfo{author}{Sch{\"a}rli, N.}, \bibinfo{author}{Hou, L.}, \bibinfo{author}{Wei, J.}, \bibinfo{author}{Scales, N.}, \bibinfo{author}{Wang, X.}, \bibinfo{author}{Schuurmans, D.}, \bibinfo{author}{Cui, C.}, \bibinfo{author}{Bousquet, O.}, \bibinfo{author}{Le, Q.~V.}, \& \bibinfo{author}{Chi, E.~H.} (\bibinfo{year}{2023}).
\newblock \bibinfo{title}{Least-to-most prompting enables complex reasoning in large language models}.
\newblock In {\it \bibinfo{booktitle}{The Eleventh International Conference on Learning Representations}\/}.

\end{thebibliography}

% %\vskip3pt

% Author biography without author photo.
% Author biography. Author biography. Author biography.
% Author biography. Author biography. Author biography.
% Author biography. Author biography. Author biography.
% Author biography. Author biography. Author biography.
% Author biography. Author biography. Author biography.
% Author biography. Author biography. Author biography.
% Author biography. Author biography. Author biography.
% Author biography. Author biography. Author biography.
% Author biography. Author biography. Author biography.

% \bio{figs/pic1}
% Author biography with author photo.
% Author biography. Author biography. Author biography.
% Author biography. Author biography. Author biography.
% Author biography. Author biography. Author biography.
% Author biography. Author biography. Author biography.
% Author biography. Author biography. Author biography.
% Author biography. Author biography. Author biography.
% Author biography. Author biography. Author biography.
% Author biography. Author biography. Author biography.
% Author biography. Author biography. Author biography.

% \bio{figs/pic1}
% Author biography with author photo.
% Author biography. Author biography. Author biography.
% Author biography. Author biography. Author biography.
% Author biography. Author biography. Author biography.
% Author biography. Author biography. Author biography.

\end{document}